\documentclass[letterpaper]{article} % DO NOT CHANGE THIS
\usepackage{aaai23}  % DO NOT CHANGE THIS
\usepackage{times}  % DO NOT CHANGE THIS
\usepackage{helvet}  % DO NOT CHANGE THIS
\usepackage{courier}  % DO NOT CHANGE THIS
\usepackage[hyphens]{url}  % DO NOT CHANGE THIS
\usepackage{graphicx} % DO NOT CHANGE THIS
\usepackage{natbib}  % DO NOT CHANGE THIS AND DO NOT ADD ANY OPTIONS TO IT
\usepackage{caption} % DO NOT CHANGE THIS AND DO NOT ADD ANY OPTIONS TO IT
\usepackage{algorithm}
\usepackage{algorithmic}

\usepackage{threeparttable}
\usepackage{booktabs}
\usepackage{enumerate}
\usepackage{multirow}
\usepackage{amsmath}
\usepackage{amssymb}

\usepackage[breaklinks=true,bookmarks=false]{hyperref}

\usepackage{newfloat}
\usepackage{listings}
\DeclareCaptionStyle{ruled}{labelfont=normalfont,labelsep=colon,strut=off} % DO NOT CHANGE THIS
\floatstyle{ruled}
\newfloat{listing}{tb}{lst}{}
\floatname{listing}{Listing}
\title{MicroAST: Towards Super-Fast Ultra-Resolution Arbitrary Style Transfer}

\author{
	Zhizhong Wang, \hspace{0.4cm}Lei Zhao\thanks{Corresponding authors.}, \hspace{0.4cm} Zhiwen Zuo, \hspace{0.4cm} Ailin Li, \\
	Haibo Chen, \hspace{0.4cm} Wei Xing$^*$, \hspace{0.4cm}Dongming Lu
}
\affiliations{
	College of Computer Science and Technology, Zhejiang University\\
	\{endywon, cszhl, zzwcs, liailin, cshbchen, wxing, ldm\}@zju.edu.cn

}

\usepackage{bibentry}
\begin{document}
	
\twocolumn[{%
	\renewcommand\twocolumn[1][]{#1}%
	\maketitle
	\vspace{-1.4cm}
	\begin{center}
		\centering
		\includegraphics[width=1\linewidth, height=0.52\linewidth]{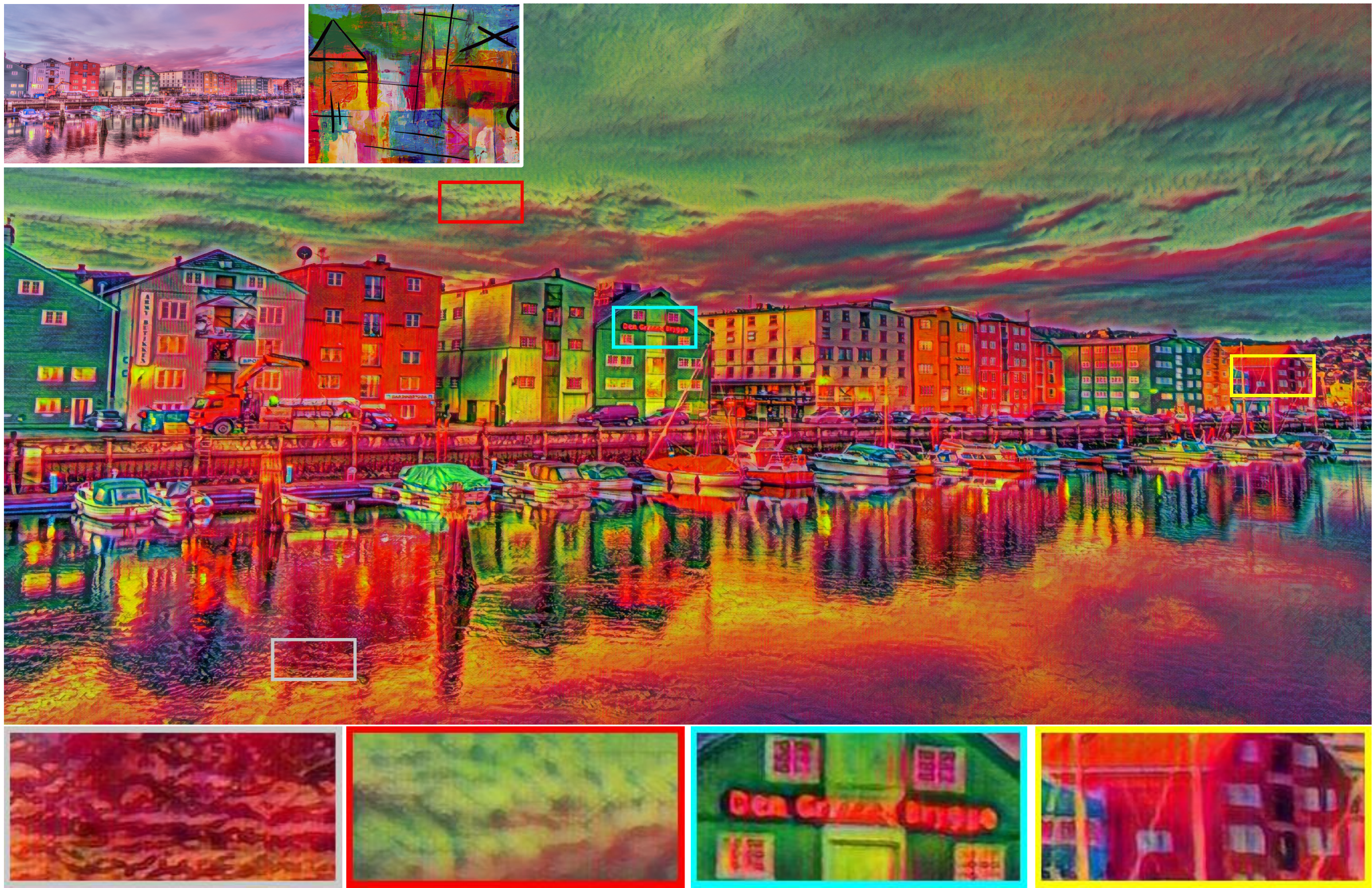}
		\vspace{-1.4em}
		\captionof{figure}{{\bf A 4K (4096$\times$2160 pixels) ultra-resolution stylized result}, rendered in about 0.5 seconds by our proposed MicroAST on a single NVIDIA RTX 2080 (8GB) GPU. On the upper left are the content and style images. Four close-ups (256$\times$128 pixels) are shown under the stylized image.
		}
		\label{fig:teaser}
		
	\end{center} 
}]

%\maketitle

\renewcommand{\thefootnote}{}
\footnote{*Corresponding authors.}
\footnote{Copyright \copyright~2023, Association for the Advancement of Artificial Intelligence (www.aaai.org). All rights reserved.}

\renewcommand{\thefootnote}{1}

\begin{abstract}
Arbitrary style transfer (AST) transfers arbitrary artistic styles onto content images. Despite the recent rapid progress, existing AST methods are either incapable or too slow to run at ultra-resolutions ({\em e.g.}, 4K) with limited resources, which heavily hinders their further applications. In this paper, we tackle this dilemma by learning a straightforward and lightweight model, dubbed {\em MicroAST}. The key insight is to completely abandon the use of cumbersome pre-trained Deep Convolutional Neural Networks ({\em e.g.}, VGG) {\em at inference}. Instead, we design two micro encoders (content and style encoders) and one micro decoder for style transfer. The content encoder aims at extracting the main structure of the content image. The style encoder, coupled with a modulator, encodes the style image into learnable {\em dual-modulation} signals that modulate both intermediate features and convolutional filters of the decoder, thus injecting more sophisticated and flexible style signals to guide the stylizations. In addition, to boost the ability of the style encoder to extract more distinct and representative style signals, we also introduce a new {\em style signal contrastive loss} in our model. Compared to the state of the art, our MicroAST not only produces visually superior results but also is 5-73 times smaller and 6-18 times faster, for the first time enabling {\em super-fast} (about 0.5 seconds) AST at 4K ultra-resolutions. Code is available at \url{https://github.com/EndyWon/MicroAST}.
\end{abstract}

\iffalse
\begin{figure}[t]
	\centering
	\includegraphics[width=1\linewidth, height=0.54\linewidth]{figs/teaser.pdf}
	\caption{{\bf A 4K (4096$\times$2160 pixels) ultra-resolution stylized result}, rendered in about 0.5 seconds by our proposed MicroAST on a single NVIDIA RTX 2080 (8GB) GPU. On the upper left are the content and style images. Four close-ups (256$\times$128 pixels) are shown under the stylized image. More results can be found in {\em supplementary material (SM)}.
	}
	\label{fig:teaser}
\end{figure}
\fi

\section{Introduction}
\label{introduction}
Style transfer has recently attracted ever-growing interest in both academia and industry since the seminal work of \cite{gatys2016image}. A central problem in this domain is the task of transferring the artistic style of an arbitrary image onto a content target, which is called {\em arbitrary style transfer (AST)}~\cite{huang2017arbitrary,li2017universal}. By leveraging the remarkable representative power of pre-trained Deep Convolutional Neural Networks (DCNNs) ({\em e.g.}, VGG-19~\cite{simonyan2014very}), existing AST algorithms consistently achieve both stunning stylizations and generalization ability on arbitrary images. However, the large pre-trained DCNNs incur a high computational cost, which impedes the current AST methods to process ultra-high resolution ({\em e.g.}, ``4K" or  4096$\times$2160 pixels) images with limited resources. It heavily restricts their further applications in practical scenes, such as large posters, ultra high-definition (UHD) photographs, and UHD videos.

Valuable efforts have been devoted to solving this dilemma. One practice is to compress the large pre-trained DCNN models without losing much performance. \cite{wang2020collaborative} used collaborative distillation to reduce the convolutional filters of VGG-19, successfully rendering ultra-resolution images on a single 12GB GPU. While the memory consumption is significantly reduced, the pruned models are often not fast enough to run at ultra-resolutions. Another solution is to stylize the images in a patch-wise manner~\cite{chen2022towards}. This method, though achieving unconstrained resolution style transfer, still suffers from the efficiency problem. Similar to our method, \cite{shen2018neural} and \cite{jing2020dynamic} likewise designed lightweight networks for style transfer. However, since their style features are still extracted from VGG, they are inherently difficult to process ultra-resolution images. Therefore, despite the recent progress, existing AST methods are still incapable or too slow to run at ultra-resolutions.

Facing the challenges above, in this paper, we propose a straightforward and lightweight model for {\em super-fast} ultra-resolution arbitrary style transfer. The key insight is that we completely abandon the use of cumbersome pre-trained DCNNs ({\em e.g.}, VGG) {\em at inference}, whether for content extraction~\cite{huang2017arbitrary,li2017universal}, or style extraction~\cite{shen2018neural,jing2020dynamic}. Our model, dubbed {\em MicroAST}, uses two micro encoders and one micro decoder for style transfer. The micro encoders consist of a content encoder and a style encoder. The content encoder aims at extracting the main structure of the content image. The style encoder, coupled with a modulator, encodes the style image into learnable {\em dual-modulation} signals that modulate both intermediate features and convolutional filters of the decoder. This novel dual-modulation strategy injects more sophisticated and flexible style signals to guide the stylizations, thus helping our model fully capture the global attributes and local brushstrokes of the artistic styles. The decoder generates the final stylized images under these modulations. In addition, to boost the ability of the style encoder to extract more distinct and representative modulation signals for each style, we also introduce a new {\em style signal contrastive loss} in our model, which further improves the quality. Comprehensive experiments have been conducted to demonstrate the effectiveness of our method. Compared to the state of the art, our MicroAST not only produces visually superior results but also is 5-73 times smaller and 6-18 times faster, for the first time enabling {\em super-fast} (about 0.5 seconds) AST on 4K ultra-resolution images (see an example in Fig.~\ref{fig:teaser}). 

In summary, our contributions are fourfold:

\begin{itemize}

	\item We propose a straightforward and lightweight framework called {\em MicroAST} to achieve {\em super-fast} ultra-resolution arbitrary style transfer for the first time.
	
	\item We introduce a novel {\em dual-modulation} strategy to inject more sophisticated and flexible style signals to guide the stylizations in our model.
	
	\item We also introduce a new {\em style signal contrastive loss} to boost the ability of our style encoder.
	
	\item Extensive qualitative and quantitative experiments demonstrate the effectiveness and superiority of our method against the state of the art.
\end{itemize}

\begin{figure*}[t]
	\centering
	\includegraphics[width=0.9\linewidth]{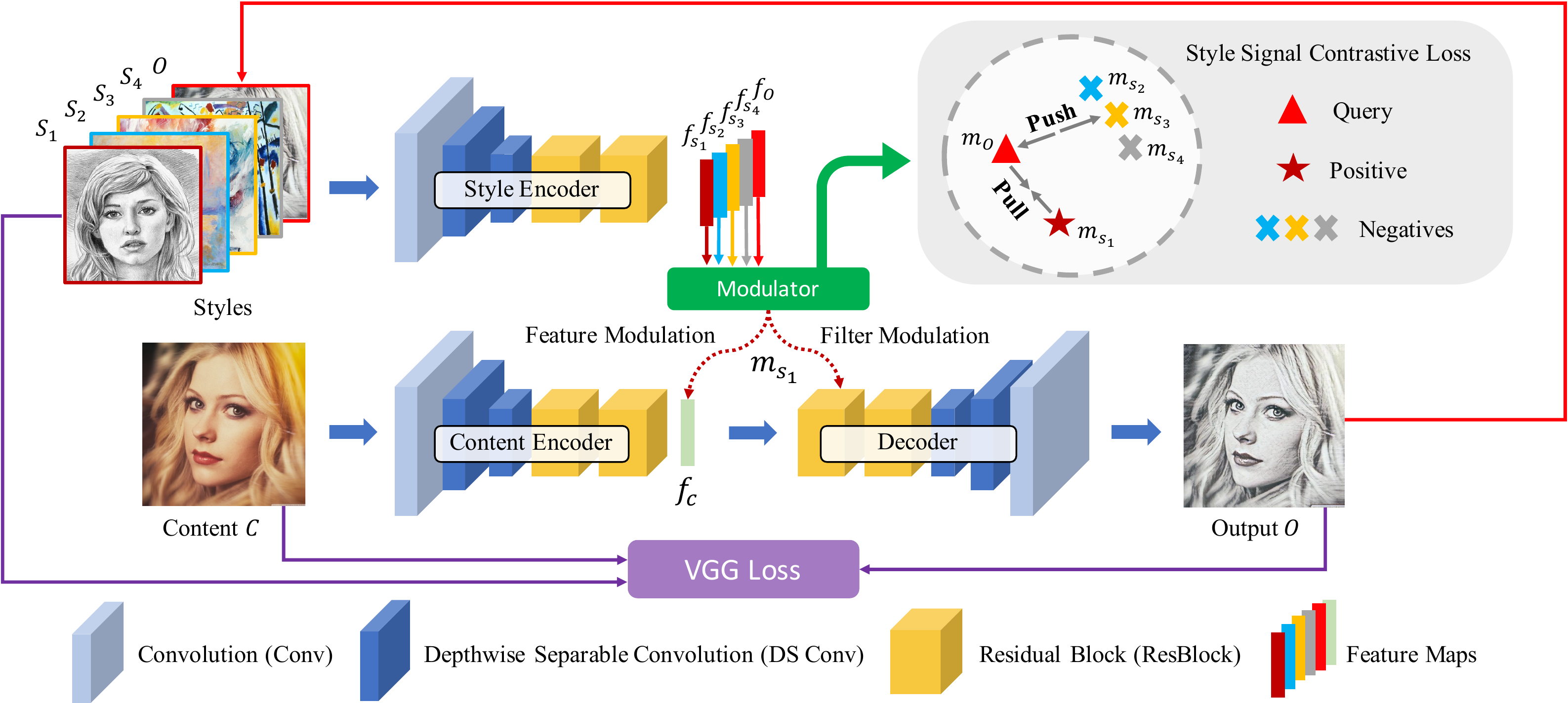}
	\caption{{\bf Overview of MicroAST.} $C$ denotes the content image. $S_i$ denotes the $i^{th}$ style image in a training mini-batch (we put four style images for illustration). $O$ denotes the stylized output. $f_{\{\dots\}}$ denote the encoded feature maps. $m_{\{\dots\}}$ denote the style modulation signals.
	}
	\label{fig:overview}
\end{figure*}

\section{Related Work}
{\bf Neural Style Transfer.} The seminal work of \cite{gatys2016image} has opened up the era of Neural Style Transfer (NST)~\cite{jing2019neural}. In their work, the artistic style of an image is captured by the correlations between features extracted from a pre-trained DCNN. It is amazingly effective and has inspired a lot of successors to improve the performance in many aspects, including efficiency~\cite{johnson2016perceptual,ulyanov2016texture}, quality~\cite{jing2018stroke,jing2022learning,wang2020glstylenet,wang2021evaluate,wang2022aesust,lin2021drafting,cheng2021style,an2021artflow,chandran2021adaptive,liu2021adaattn,chen2021dualast,chen2021artistic,deng2020arbitrary,deng2021arbitrary,deng2022stytr2,xie2022artistic,kwon2022clipstyler}, generalization~\cite{huang2017arbitrary,li2017universal,sheng2018avatar,li2019learning,park2019arbitrary,lu2019closed,zhang2019metastyle,chiu2019understanding,chiu2020iterative,hong2021domain,zhang2022exact}, diversity~\cite{wang2020diversified,wang2022divswapper,chen2020creative,chen2021diverse}, and user control~\cite{champandard2016semantic,wang2022texture,zuo2022style}. Despite the monumental progress, existing NST methods all share a fundamental flaw, {\em i.e.}, they are unable to process ultra-resolution ({\em e.g.}, 4K) images with limited resources, since they all heavily rely on the large DCNN models ({\em e.g.}, VGG-19~\cite{simonyan2014very}) to extract representative features.

{\bf Ultra-Resolution Style Transfer.} To address the challenges above, \cite{wang2020collaborative} used model compression (called collaborative distillation) to reduce the convolutional filters of VGG-19, firstly rendering ultra-resolution images on a 12GB GPU. While the memory consumption is significantly reduced, the pruned models are still not fast enough to run at ultra-resolutions. Besides, a large degree of compression often leads to severe quality degradation.

Another solution is to design a lightweight model directly. \cite{johnson2016perceptual} and \cite{sanakoyeu2018style} learned small feed-forward networks for a specific style example or category for high-resolution ({\em e.g.}, 1024$\times$1024 pixels) style transfer. However, they are not generalized to other unseen styles and not capable of running at ultra-resolutions. \cite{shen2018neural} and \cite{jing2020dynamic} designed lightweight networks for AST, but they still used pre-trained VGG to extract style features, leading to the expensive cost of extra memory and slow inference speed. Unlike these methods, our approach completely gets rid of the high-cost pre-trained VGG {\em at inference}, for the first time achieving {\em super-fast} ultra-resolution style transfer for arbitrary styles with one model only. 

Recently, \cite{chen2022towards} provided a possible solution for unconstrained resolution style transfer. They divided input images into small patches and performed patch-wise stylization with a Thumbnail Instance Normalization to ensure the style consistency among different patches. However, this method does not consider the time cost problem and cannot achieve {\em super-fast} ultra-resolution style transfer.

{\bf Contrastive Learning.} Contrastive learning has been widely used in self-supervised representation learning for high-level vision tasks~\cite{he2020momentum,chen2020simple,tian2020contrastive}. Recently, in low-level generative tasks, some works investigate the use of contrastive loss for different objectives, such as image-to-image translation~\cite{park2020contrastive}, image generation~\cite{kang2020contragan,liu2021divco}, image dehazing~\cite{wu2021contrastive}, and style transfer~\cite{chen2021artistic,zhang2022domain,wu2022ccpl},~{\em etc}. Unlike these works, we introduce a novel mini-batch {\em style signal contrastive loss} to help address the problem of ultra-resolution style transfer, which considers relations between multiple style modulation signals in the same training mini-batches. Therefore, it can significantly boost the ability of the micro style encoder to extract more distinct and representative style signals.

\section{Proposed Approach}
\label{approach}
Given {\em arbitrary} ultra-resolution ({\em e.g.}, 4K) content image $C$ and style image $S$, our goal is to produce the corresponding ultra-resolution stylized output $O$ in a {\em very short} time (e.g., within one second). The challenges mainly lie in three aspects. (i) The method should be capable of processing ultra-resolution images with {\em limited resources} (e.g., on an 8GB GPU). (ii) The method should generate the ultra-resolution results at a {\em super-fast} speed. (iii) The method should be able to produce pleasing stylizations for {\em arbitrary} contents and styles. To achieve these goals, we propose a novel {\em MicroAST} framework, which will be introduced in detail.

\renewcommand\arraystretch{0.6}
\begin{figure*}[htbp]
	\centering
	\setlength{\tabcolsep}{0.15cm}
	\begin{tabular}{c|c|c|c}
		\includegraphics[width=0.235\linewidth]{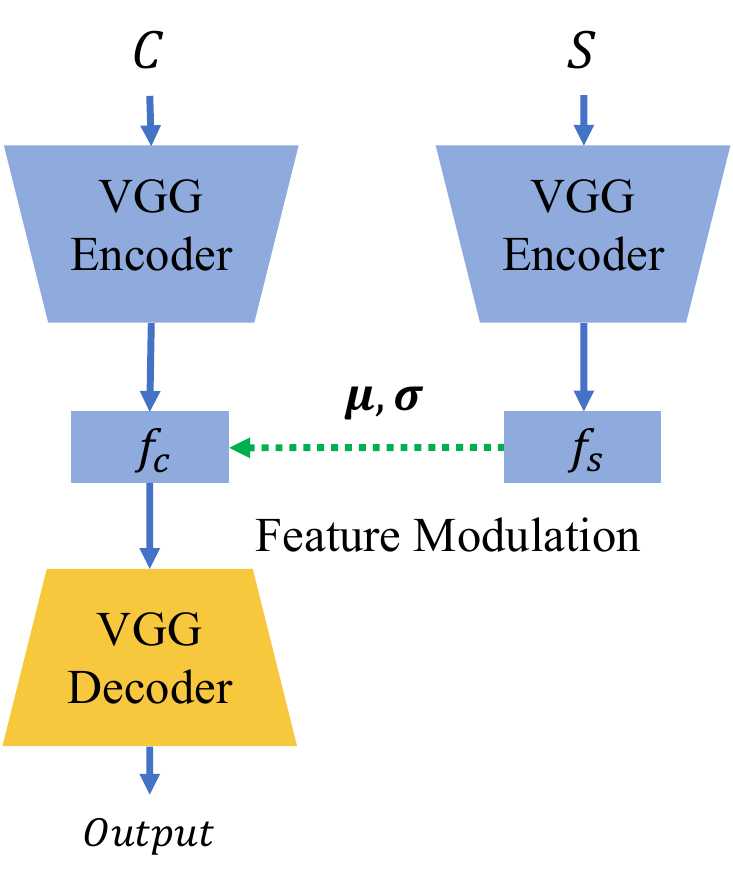}&
		\includegraphics[width=0.235\linewidth]{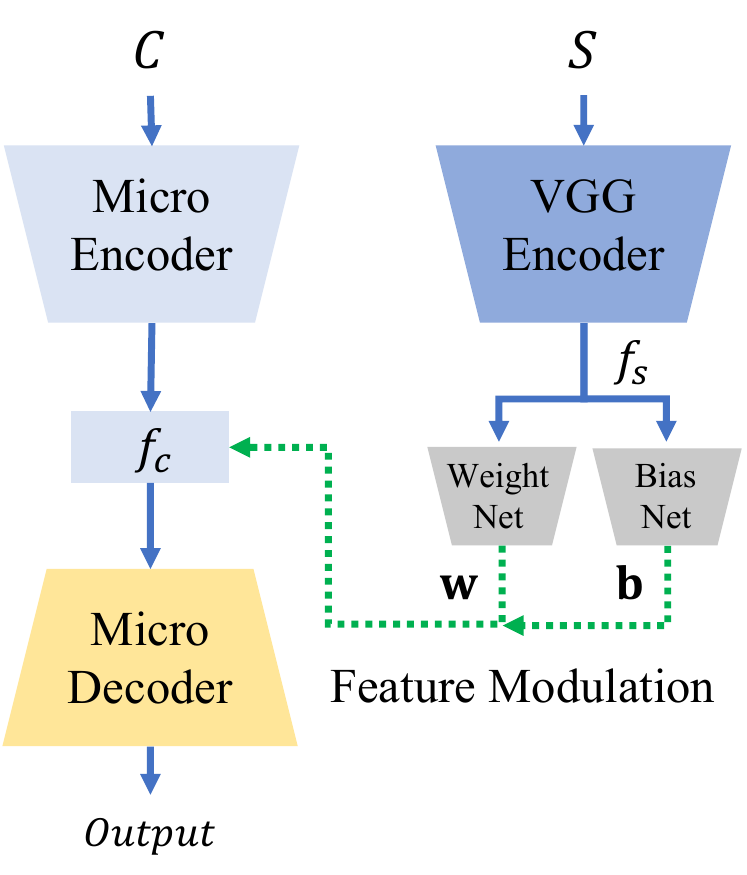}&
		\includegraphics[width=0.235\linewidth]{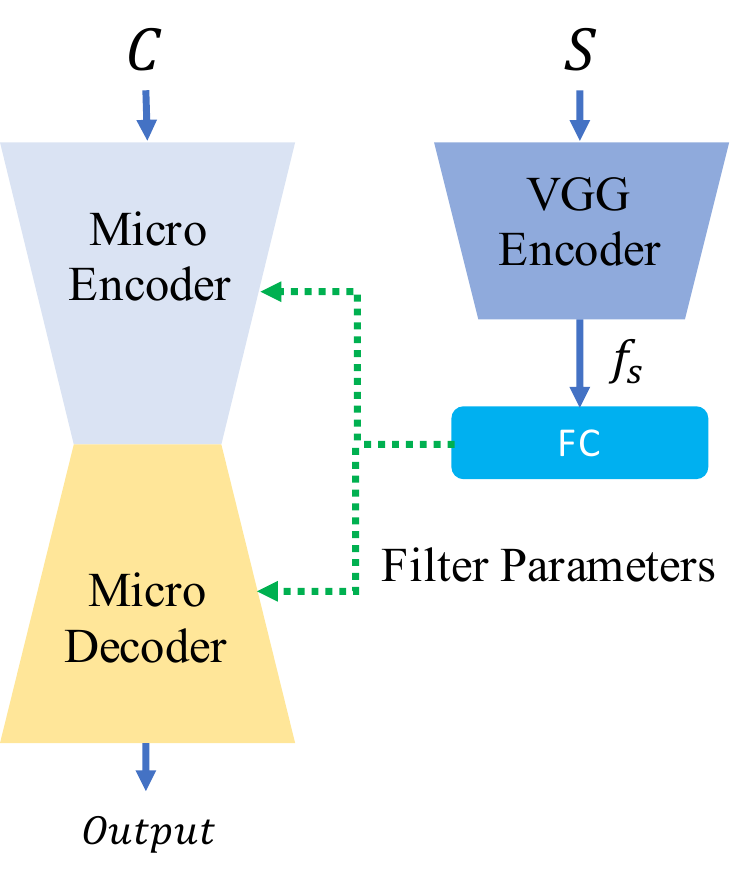}&
		\includegraphics[width=0.235\linewidth]{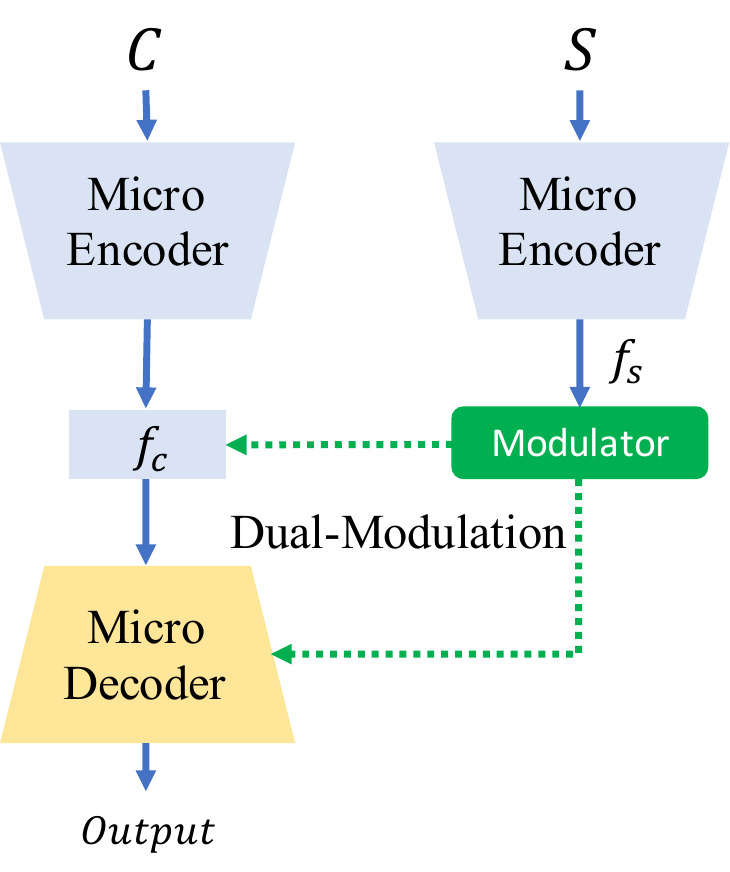}
		\\
		\footnotesize (a) AdaIN & \footnotesize (b) DIN & \footnotesize (c) MetaNets &  \footnotesize \bf (d) Ours
	\end{tabular}
	\caption{{\bf Illustration from modulation to dual-modulation.} Existing AST methods either directly modulate/normalize the intermediate features (a, b), or  straightforwardly generate the parameters of network filters (c). Our method modulates both intermediate features and network filters (d). ``FC'' stands for fully connected layers. The details of our dual-modulation are illustrated in Fig.~\ref{fig:mod}.}
	\label{fig:mods}
\end{figure*}

\subsection{Overview of MicroAST}
As illustrated in Fig.~\ref{fig:overview}, our MicroAST consists of three main components: a micro content encoder $E_c$, a micro style encoder $E_s$ (coupled with a modulator $\mathcal{M}$), and a micro decoder $D$. In details, $E_c$ and $E_s$ have the same lightweight architecture which comprises 1 standard stride-1 convolutional (Conv) layer, 2 stride-2 depthwise separable convolutional (DS Conv) layers, and 2 stride-1 residual blocks (ResBlocks). The micro decoder $D$ is mostly symmetrical to the encoders. The modulator $\mathcal{M}$ consists of two subnets as shown in Fig.~\ref{fig:mod} (see later Sec.~\ref{DM}). More detailed architectures can be found in {\em supplementary material (SM)}\footnote{\url{https://github.com/EndyWon/MicroAST/releases/download/v1.0.0/MicroAST\_SM.pdf}}. Note that since our model is based on MobileNet~\cite{howard2017mobilenets}, it can be easily applied to mobile devices. The overall pipeline is as follows:
\begin{enumerate}
	\renewcommand{\labelenumi}{(\theenumi)}
	\item Extract the main structure of the content image $C$ using the micro content encoder $E_c$, denoted as $f_c:=E_c(C)$.
	
	\item Extract the style feature of the style image $S$ using the micro style encoder $E_s$, denoted as $f_s:=E_s(S)$.
	
	\item Convert the style feature $f_s$ into the style modulation signals (a set of learnable parameters) using the modulator $\mathcal{M}$, denoted as $m_s := \mathcal{M}(f_s)$.
	
	\item Stylize $f_c$ using the micro decoder $D$, under the dual-modulations (Sec.~\ref{DM}) of $m_s$, {\em i.e.}, $O:=D(f_c, m_s)$.
\end{enumerate}

{\bf Training Losses.} To achieve style transfer, similar to previous works~\cite{gatys2016image,johnson2016perceptual,huang2017arbitrary}, we leverage a pre-trained VGG-19~\cite{simonyan2014very} as our loss network to compute the content loss and style loss. We use the perceptual loss~\cite{johnson2016perceptual} as our content loss $\mathcal{L}_c$, which is computed at layer $\{relu4\_1\}$ of VGG-19. The style loss $\mathcal{L}_s$ is defined to match the Instance Normalization (IN) statistics~\cite{huang2017arbitrary}, which is computed at layer $\{relu1\_1, relu2\_1, relu3\_1, relu4\_1\}$. {\em Note that the VGG-19 is only used in our training phase, and our model does not involve any large network at inference.}

To further improve the stylization quality, we also introduce a novel {\em style signal contrastive (SSC) loss} $\mathcal{L}_{ssc}$ to train our model. This loss could help boost the ability of the micro style encoder to extract more distinct and representative modulation signals for each style (see details in Sec.~\ref{SSCL}).

To summarize, the full objective of our MicroAST is:
\begin{equation}
	\label{eq:loss}
	\mathcal{L}_{full} := \lambda_c\mathcal{L}_c + \lambda_s\mathcal{L}_s + \lambda_{ssc}\mathcal{L}_{ssc},
\end{equation}
where hyper-parameters $\lambda_c$, $\lambda_s$, and $\lambda_{ssc}$ define the relative importance of the components in the overall
loss function.

\subsection{Dual-Modulation}
\label{DM}

{\bf Revisiting Modulation Strategies in AST}

{\bf (1) AdaIN.} \cite{huang2017arbitrary} first provided a generic modulation strategy for AST, namely Adaptive Instance Normalization (AdaIN). As illustrated in Fig.~\ref{fig:mods} (a), AdaIN modulates the content feature $f_c$ with the channel-wise mean $\mu(\cdot)$ and standard deviation $\sigma(\cdot)$ of the style feature~$f_s$.
\begin{equation}
	AdaIN(f_c, f_s) := \sigma(f_s) (\frac{f_c - \mu(f_c)}{\sigma(f_c)}) + \mu(f_s).
\end{equation}
While AdaIN has obtained great success in recent generative models~\cite{karras2019style,karras2020analyzing}, in style transfer, \cite{jing2020dynamic} pointed out that there are two critical requirements for AdaIN. (1) The content and style encoders should be identical. (2) The network architecture of the encoders should be complex enough, like VGG. Obviously, requirement (2) is contrary to our task, and requirement (1) hinders the ability of the encoders to extract domain-specific features, especially when they are micro. Therefore, the AdaIN system is not suitable for our task.

{\bf (2) DIN.} To address the requirement (1) of AdaIN and also train a lightweight network, \cite{jing2020dynamic} proposed Dynamic Instance Normalization (DIN). As illustrated in Fig.~\ref{fig:mods}~(b), DIN uses a micro content encoder to extract the content feature $f_c$. The style feature $f_s$ is extracted from a pre-trained sophisticated VGG encoder, along with two subnets to generate the dynamic normalization weight $\bf w$ and bias $\bf b$.
\begin{equation}
	DIN(f_c, f_s) := {\bf w} (\frac{f_c - \mu(f_c)}{\sigma(f_c)}) + {\bf b}.
\end{equation}
These learned dynamic parameters lead to a more accurate alignment of the real complex statistics of style features~\cite{jing2020dynamic}, and the micro content encoder and decoder drastically reduce the model size. However, since it still uses the high-cost VGG encoder to extract the style features, the DIN system also cannot be adopted in our task.

{\bf (3) MetaNets.} Like DIN, \cite{shen2018neural} also trained a lightweight network for AST. The difference is that they modulate the networks directly instead of the features. As illustrated in Fig.~\ref{fig:mods} (c), the style image $S$ is first fed into the fixed pre-trained VGG to get the style feature $f_s$, and then goes through fully connected (FC) layers to construct the filters for each $Conv$ layer in the corresponding image transformation network. The VGG and the FC layers are called ``MetaNets''. While they can convert an arbitrary new style into a lightweight image transformation network, the high-cost VGG and FC layers lead to the expensive cost of extra memory and slow genuine inference time. Again, the MetaNets system cannot be used for our task.

{\bf Dual-Modulation: FeatMod and FilterMod}

\begin{figure}[t]
	\centering
	\setlength{\tabcolsep}{0.02cm}
	\begin{tabular}{ccccc}
		\includegraphics[width=0.195\linewidth]{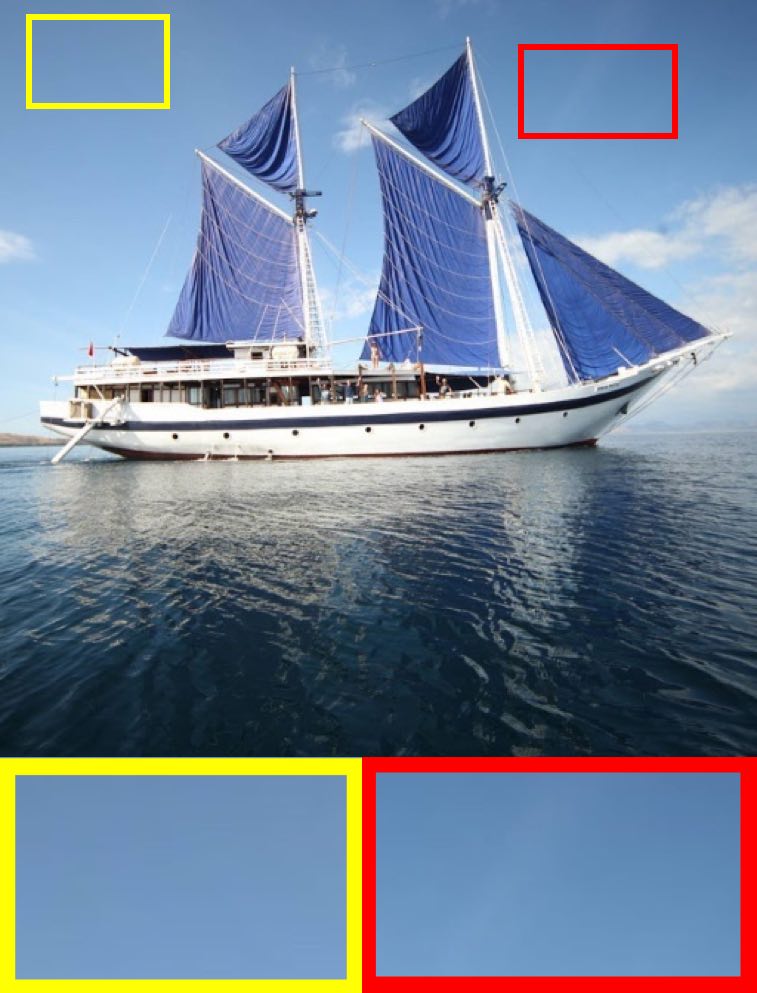}&
		\includegraphics[width=0.195\linewidth]{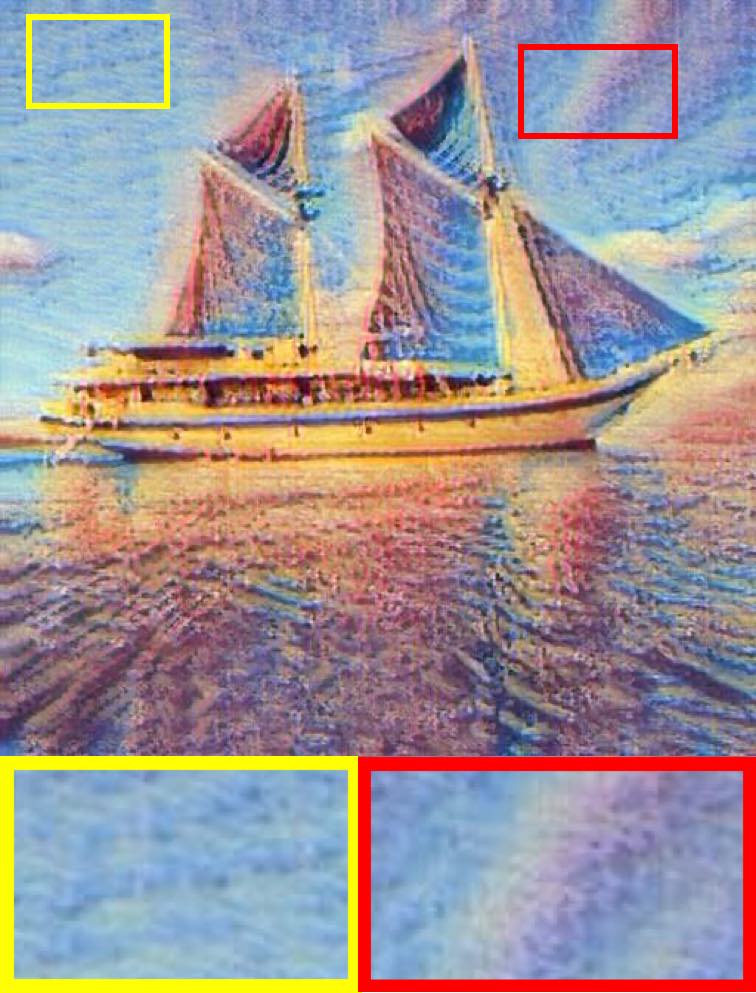}&
		\includegraphics[width=0.195\linewidth]{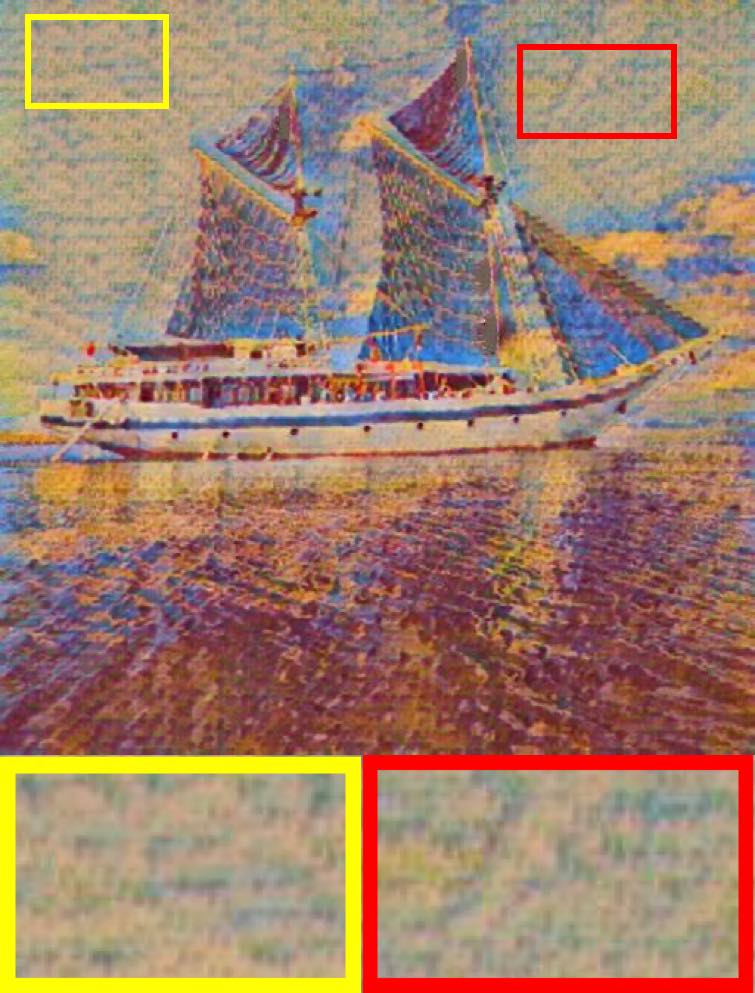}&
		\includegraphics[width=0.195\linewidth]{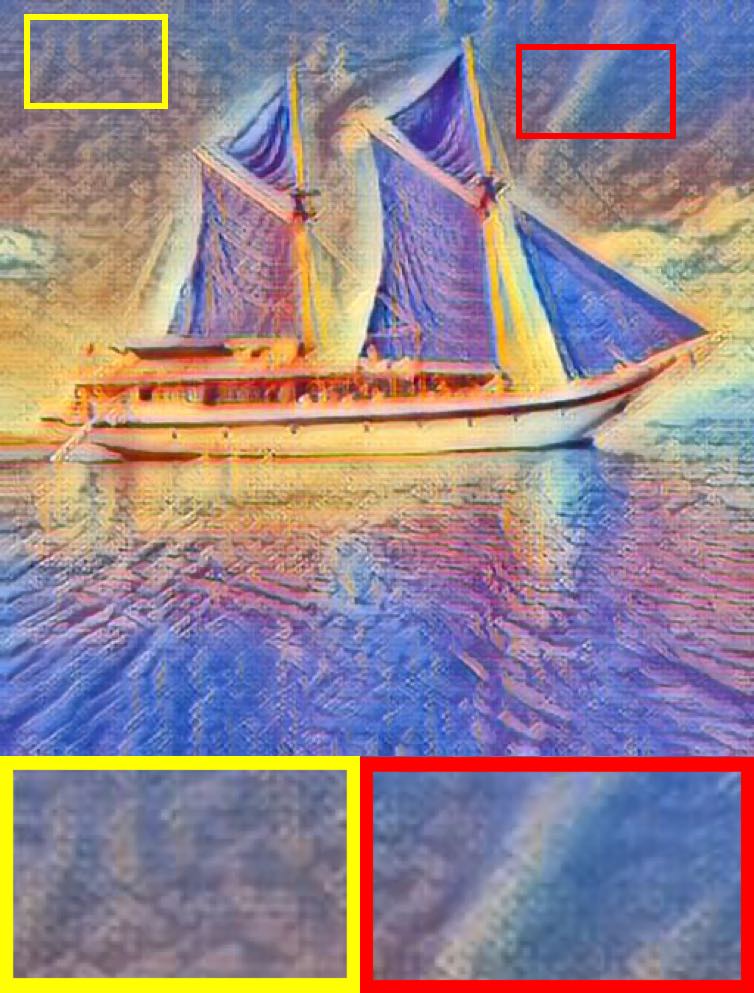}&
		\includegraphics[width=0.195\linewidth]{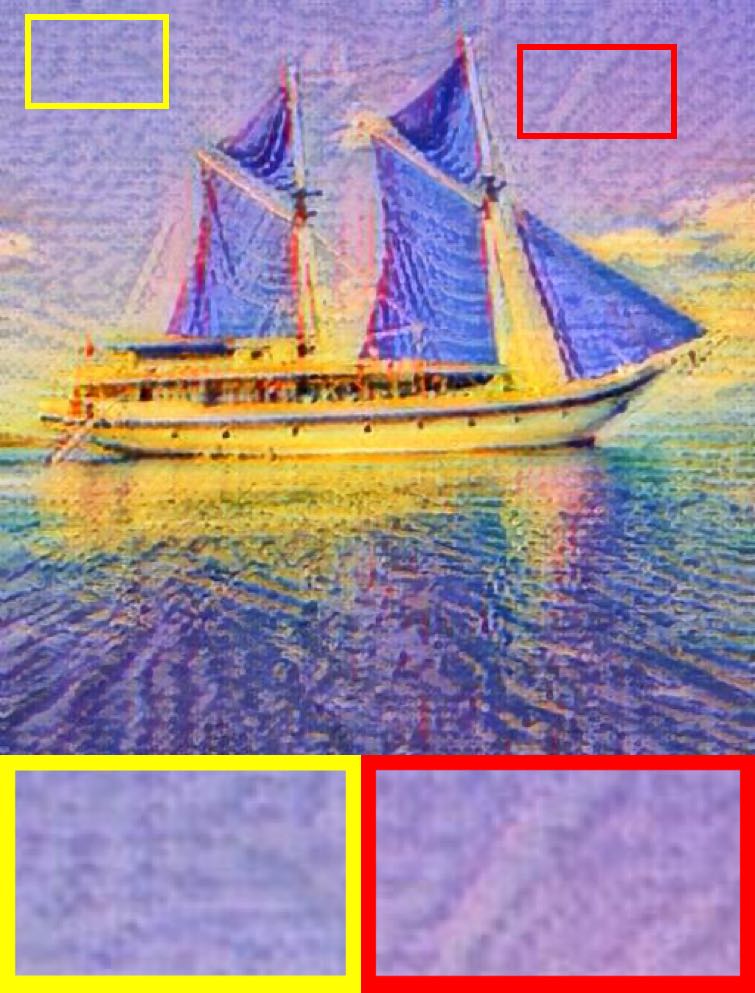}
		\\
		\scriptsize Content  & \scriptsize (a) DIN-v &  \scriptsize (b) MetaNets-v & \scriptsize (c) FilterMod-m & \scriptsize (d) DualMod-v
		\\
		\includegraphics[width=0.195\linewidth]{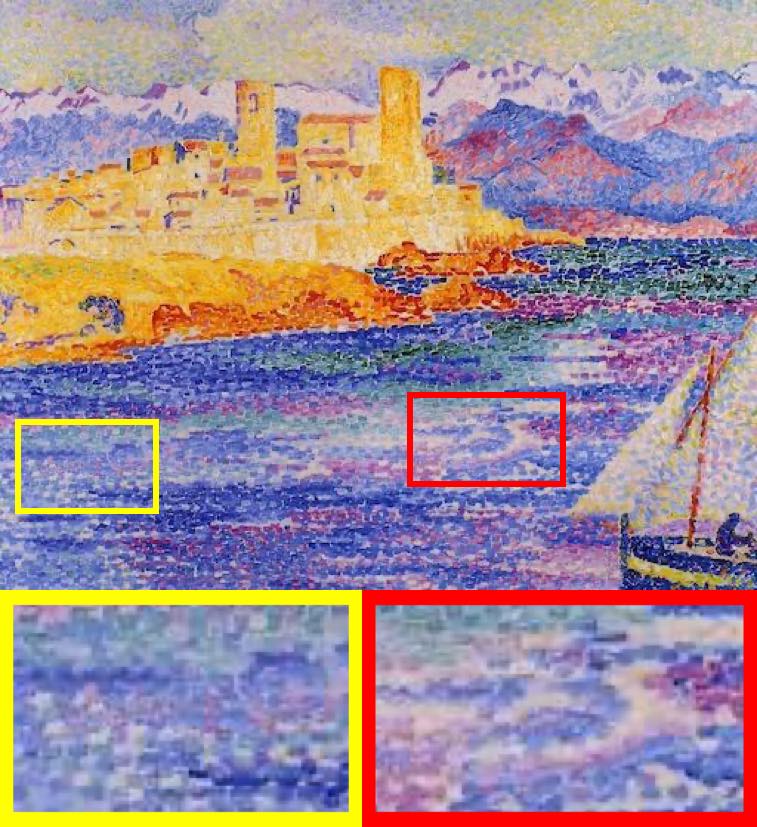}&
		\includegraphics[width=0.195\linewidth]{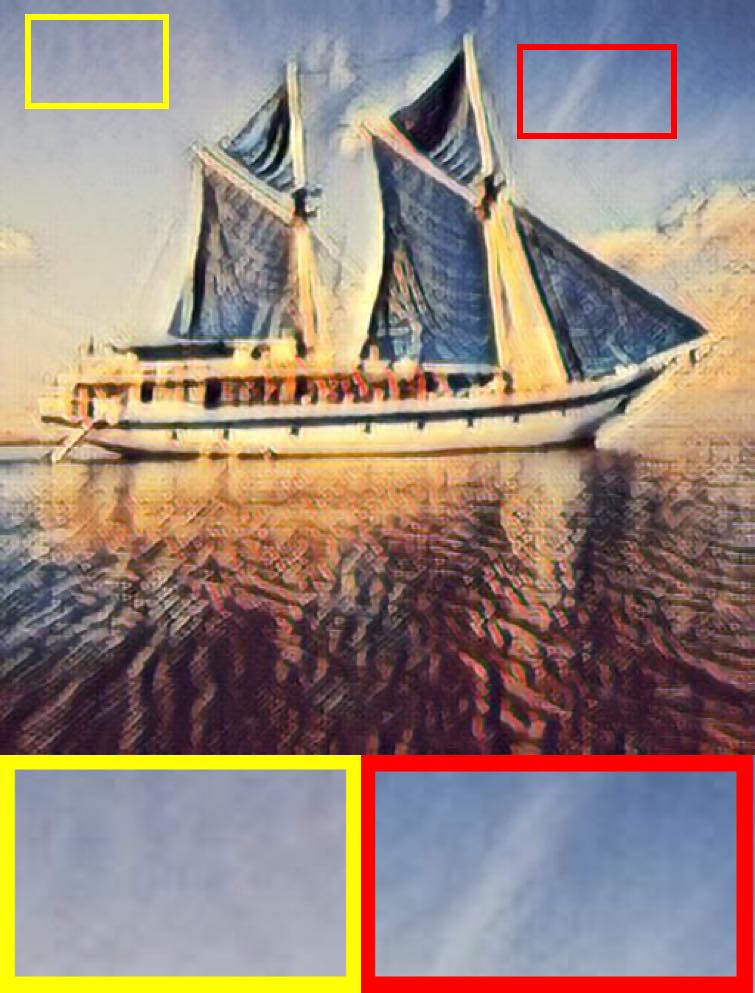}&
		\includegraphics[width=0.195\linewidth]{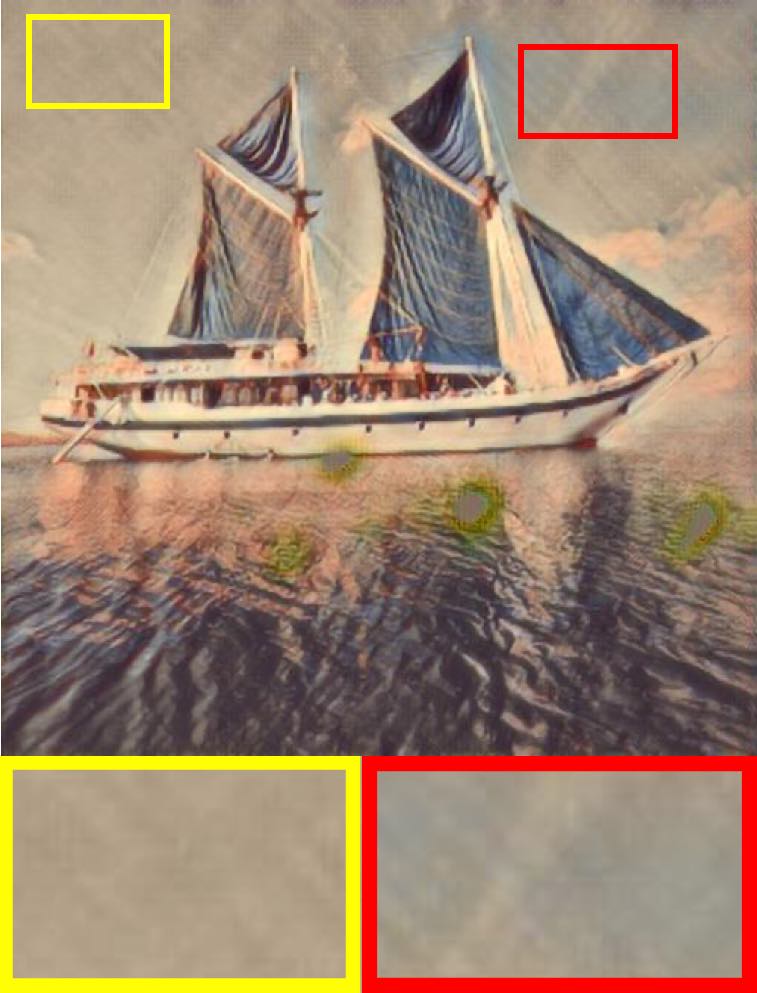}&
		\includegraphics[width=0.195\linewidth]{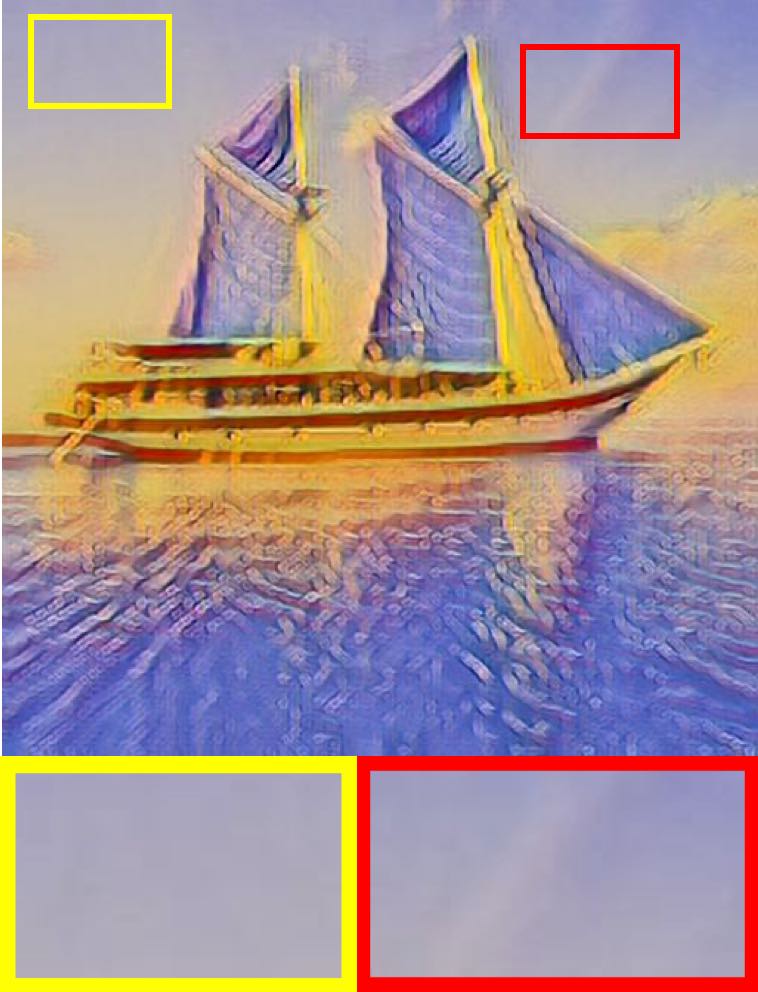}&
		\includegraphics[width=0.195\linewidth]{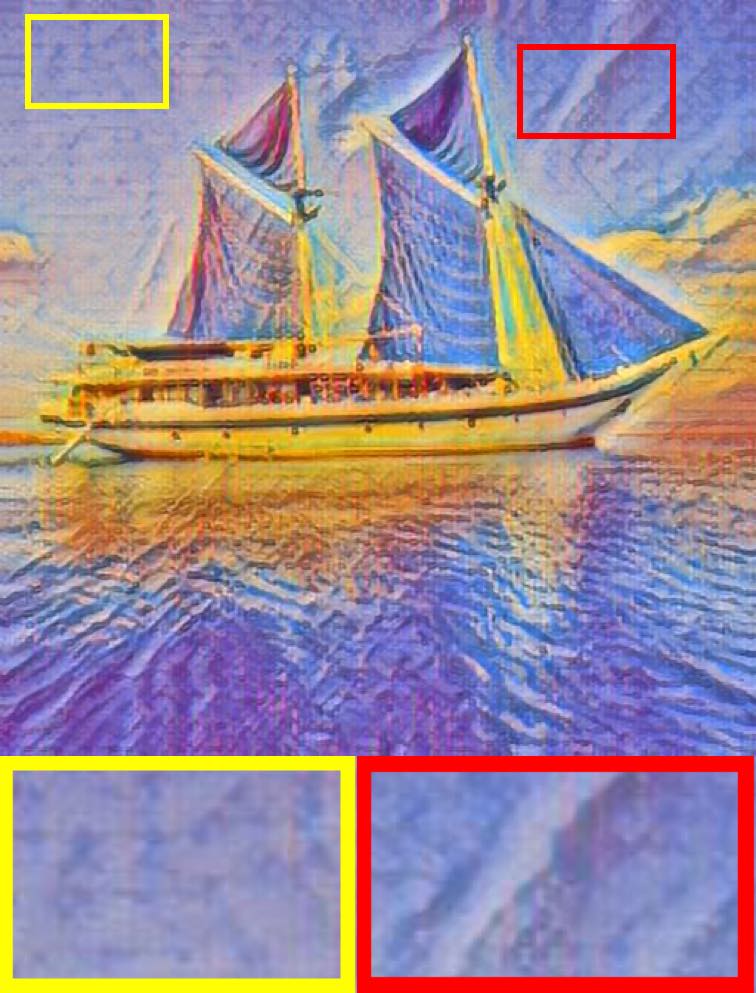}
		\\
		\scriptsize Style  & \scriptsize (e) DIN-m & \scriptsize (f) MetaNets-m & \scriptsize (g) FeatMod-m & \scriptsize (h) DualMod-m
		
	\end{tabular}
	\caption{Comparisons of VGG style encoder (marked by ``-v") vs. micro style encoder (marked by ``-m''), and modulation vs. dual-modulation (last two columns). {\em Please see SM for quantitative comparisons and more ablation studies}.
	}
	\label{fig:modill}
\end{figure}

As analyzed above, the main problem preventing DIN and MetaNets from being applied to our task is the high-cost VGG style encoder. Hence, a simple solution is to replace VGG with a micro encoder to extract the style features. Unfortunately, since the complex pre-trained VGG is also the key of these methods to achieve satisfactory stylizations, the alteration will severely degrade the quality. As shown in the $2^{nd}$ and $3^{rd}$ columns of Fig.~\ref{fig:modill}, the VGG style encoder helps DIN and MetaNets to capture complex style patterns like the punctate brushstrokes (top row, best viewed in insets below). However, when replacing VGG with a micro style encoder, these methods consistently learn few texture patterns (bottom row). We attribute it to two main factors: (1) The micro style encoder has limited ability to extract sufficiently complex style features due to the simple network architecture. (2) The style signals injected to guide the stylizations are unitary and inflexible. To address these two problems, we introduce two critical designs in our MicroAST. For the former, we propose a novel {\em contrastive loss} to boost the ability of the micro style encoder, which will be presented in later Sec.~\ref{SSCL}. For the latter, we propose an innovative {\em dual-modulation} strategy to inject more sophisticated and flexible style signals to~guide the stylizations, which will be introduced in the following.

Our dual-modulation/DualMod strategy seeks to modulate the stylization process from two different dimensions, {\em i.e.}, intermediate features (feature modulation/FeatMod) and network filters (filter modulation/FilterMod). The motivation comes up from the literature that FeatMod mainly captures the global attributes like rough textures, colors, contrast, and saturation~\cite{huang2017arbitrary}, while FilterMod is particularly good at capturing local changes like different brushstrokes~\cite{alharbi2019latent}. 

\begin{figure}[t]
	\centering
	\includegraphics[width=1\linewidth]{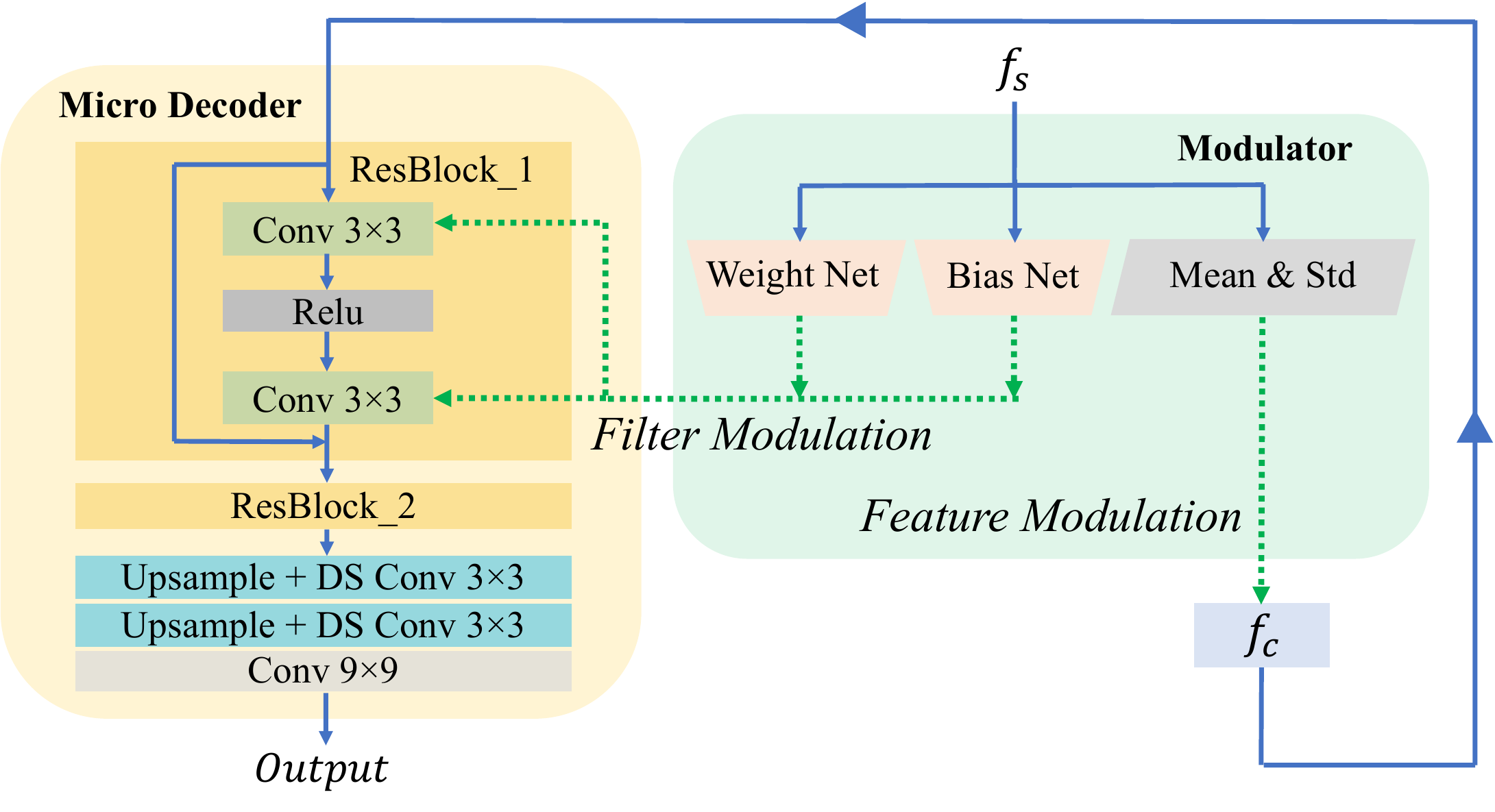}
	\caption{{\bf Details of our dual-modulation strategy.}
	}
	\label{fig:mod}
\end{figure}

{\bf FeatMod.} Concretely, our FeatMod adopts the {\em learned} channel-wise mean and standard deviation as style signals to modulate the intermediate features.
\begin{equation}
	\begin{aligned}
		&\qquad \qquad {\boldsymbol \mu_s} := \mu(f_s),  \quad  {\boldsymbol  \sigma_s} := \sigma(f_s), \\
		&FeatMod(f_c, ({\boldsymbol  \mu_s}, {\boldsymbol \sigma_s})) := {\boldsymbol \sigma_s} (\frac{f_c - \mu(f_c)}{\sigma(f_c)}) + {\boldsymbol  \mu_s}.
	\end{aligned}
\end{equation}
Note that it is different from the AdaIN system, as our style signals are {\em dynamically} learned from the trainable style encoder $E_s$, while AdaIN's are {\em statically} computed from the fixed pre-trained VGG. Also, it is unlike the DIN system, as we use the learned channel-wise mean and standard deviation as style signals, while theirs are computed by two subnets. The reason for using mean and standard deviation is that these statistics can capture the global attributes more effectively~\cite{huang2017arbitrary}, as verified in Fig.~\ref{fig:modill}~(g). By contrast, DIN learns poor on global effects like colors (see Fig.~\ref{fig:modill}~(e)), even with the VGG style encoder (see Fig.~\ref{fig:modill}~(a)).

{\bf FilterMod.} While FeatMod has been able to capture the global attributes well, it is not enough for style transfer, since the local textures like brushstrokes are also important for artistic styles~\cite{kotovenko2021rethinking}. To combat this limitation, we propose a novel FilterMod method in our model.

As illustrated in Fig.~\ref{fig:mod}, the encoded style feature $f_s$ is first converted to the weight $\bf w_s$ and bias $\bf b_s$ parameters via two simple subnets (weight net $\xi_w$ and bias net $\xi_b$ ). Then, these parameters are injected into decoder $D$ to modulate the $Conv$ filters of the ResBlocks.
\begin{equation}
	\begin{aligned}
		&\qquad \qquad  {\bf  w_s} := \xi_w(f_s), \quad {\bf  b_s} := \xi_b(f_s), \\
		&FilterMod(D, ({\bf w_s}, {\bf b_s})) := ResBlock(f_c, ({\bf w_s}, {\bf b_s})) \\
		& := Conv(Relu(Conv(f_c, ({\bf w_s}, {\bf b_s}))), ({\bf w_s}, {\bf b_s})) + f_c.
	\end{aligned}
\end{equation}
We modulate ResBlocks since they occupy the main complexity of the decoder and dominate the style transfer process~\cite{johnson2016perceptual}.

Moreover, in many deep learning platforms, it is easier to handle feature maps than filters. Therefore, by the distributive and associative property of convolution, we deduce our FilterMod to an equivalent pseudo FeatMod form as follows:
\begin{equation}
	\begin{aligned}
		Conv(f_c, ({\bf w_s}, {\bf b_s})) &:= ({\bf w_s} * \mathcal{F} + {\bf b_s}) \circledast f_c \\
		&:= ({\bf w_s} * \mathcal{F}) \circledast f_c + {\bf b_s} \circledast f_c\\
		&:= {\bf w_s} * (\mathcal{F} \circledast f_c) + {\bf  b_s} * f_c,
	\end{aligned}
\end{equation}
where $\mathcal{F}$ denotes the convolutional filter, $*$ is the element-wise multiplication with broadcast over the spatial dimensions, and $\circledast$ stands for convolution.

As shown in Fig.~\ref{fig:modill} (c), under the FilterMod, our MicroAST can capture the challenging punctate brushstrokes well even with the micro style encoder, but the global attributes like colors and contrast are not so good as FeatMod.

{\bf DualMod.} Finally, our DualMod is a combination of FeatMod and FilterMod so as to absorb both their merits.
\begin{equation}
	\begin{aligned}
		m_s := (& {\boldsymbol \mu_s}, {\boldsymbol \sigma_s}, {\bf  w_s}, {\bf  b_s}), \\
		DualMod(D, f_c, m_s) := & FeatMod(f_c, ({\boldsymbol \mu_s}, {\boldsymbol \sigma_s})) + \\
		&FilterMod(D, ({\bf  w_s}, {\bf  b_s})), 
	\end{aligned}
	\label{eq:dm}
\end{equation}
where $m_s$ are the style modulation signals.

As validated in Fig.~\ref{fig:modill} (h), DualMod helps our MicroAST capture both global attributes and local brushstrokes well. The result is very encouraging and almost on a par with that produced by using VGG style encoder (see Fig.~\ref{fig:modill} (d)).

\renewcommand\arraystretch{0.1}
\begin{figure*}[t]
	\centering
	\setlength{\tabcolsep}{0.04cm}
	\begin{tabular}{cccccc}
		\includegraphics[width=0.0851\linewidth]{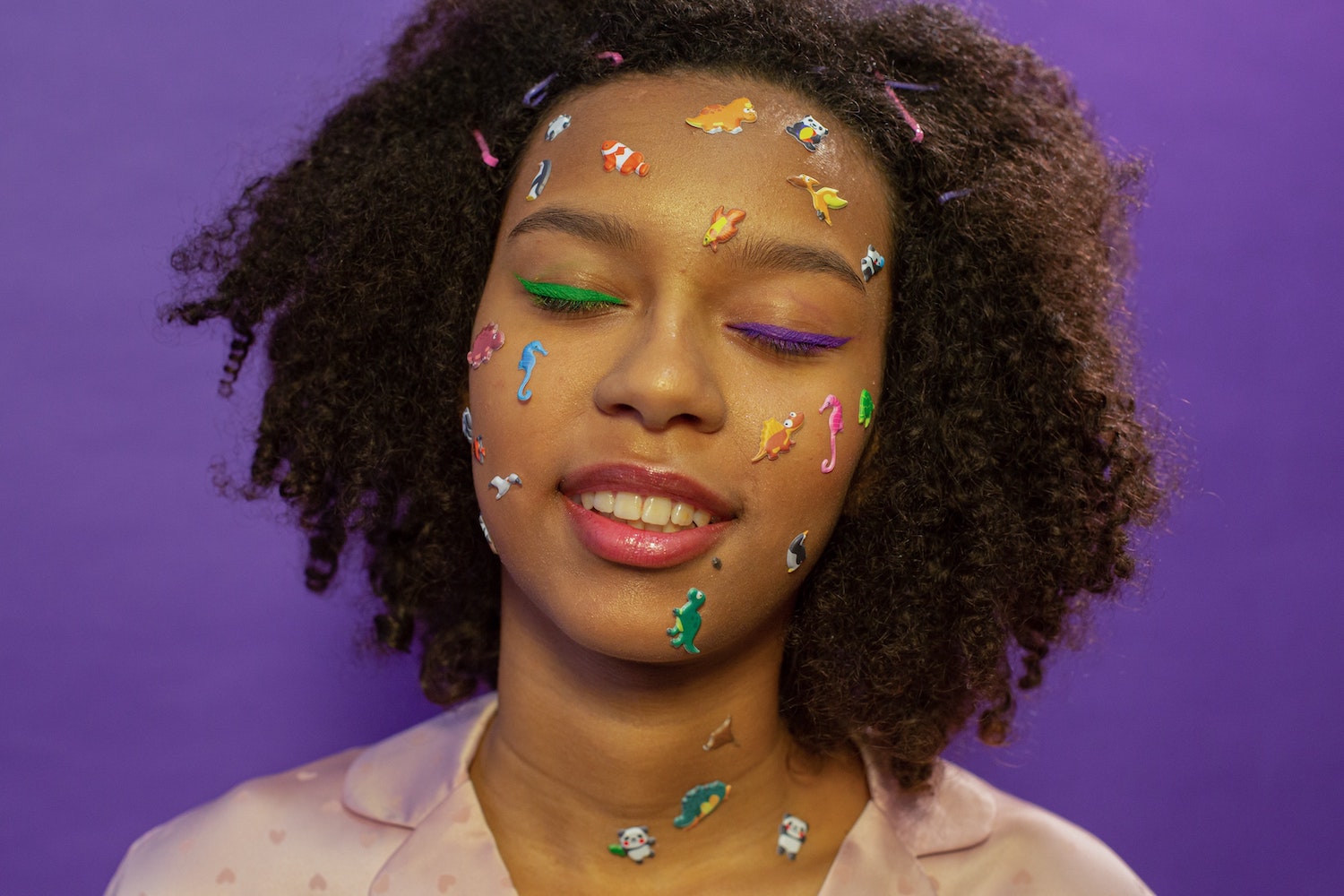}&
		\multirow{2}{*}[0.386in]{\includegraphics[width=0.171\linewidth]{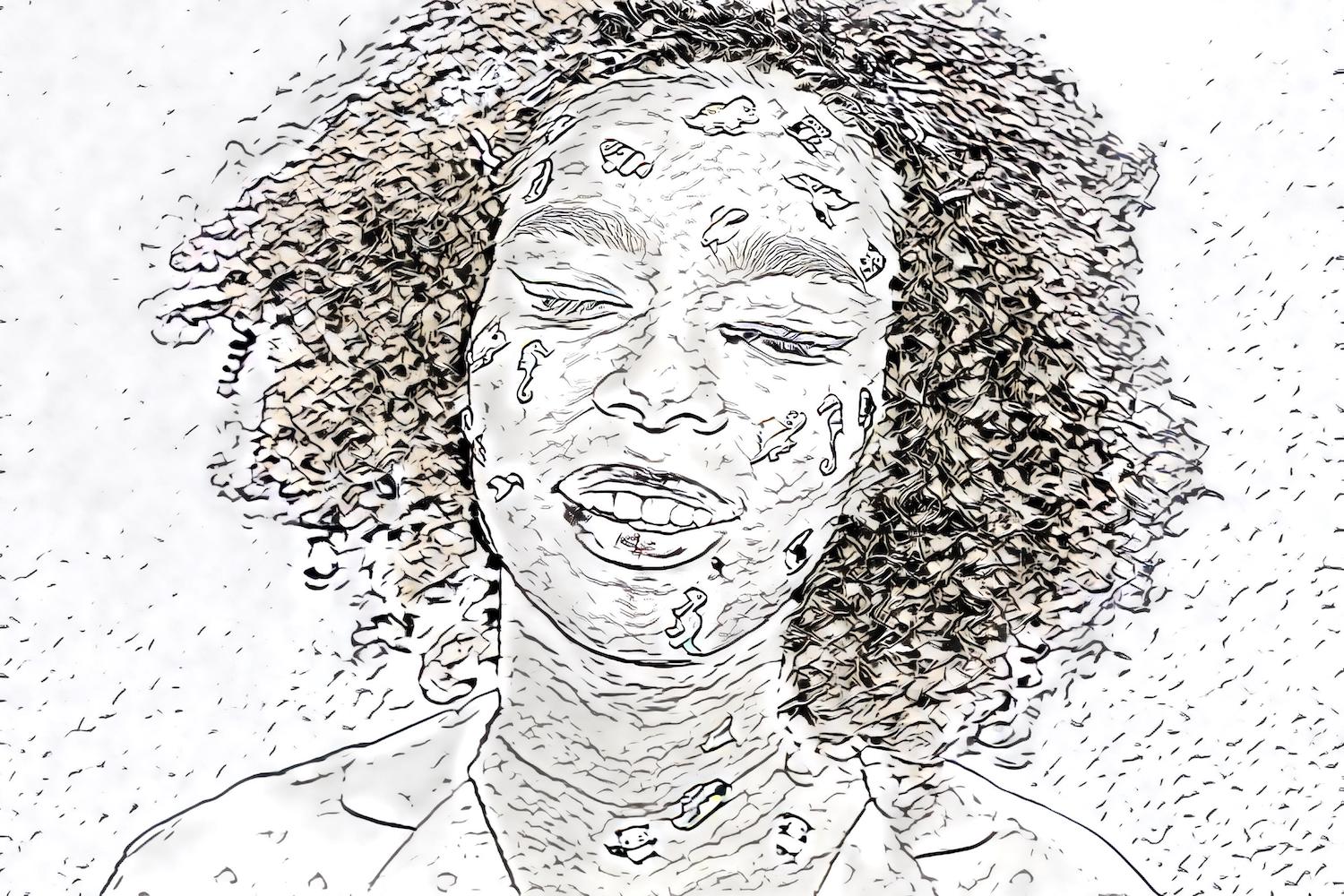}}&
		\multirow{2}{*}[0.386in]{\includegraphics[width=0.171\linewidth]{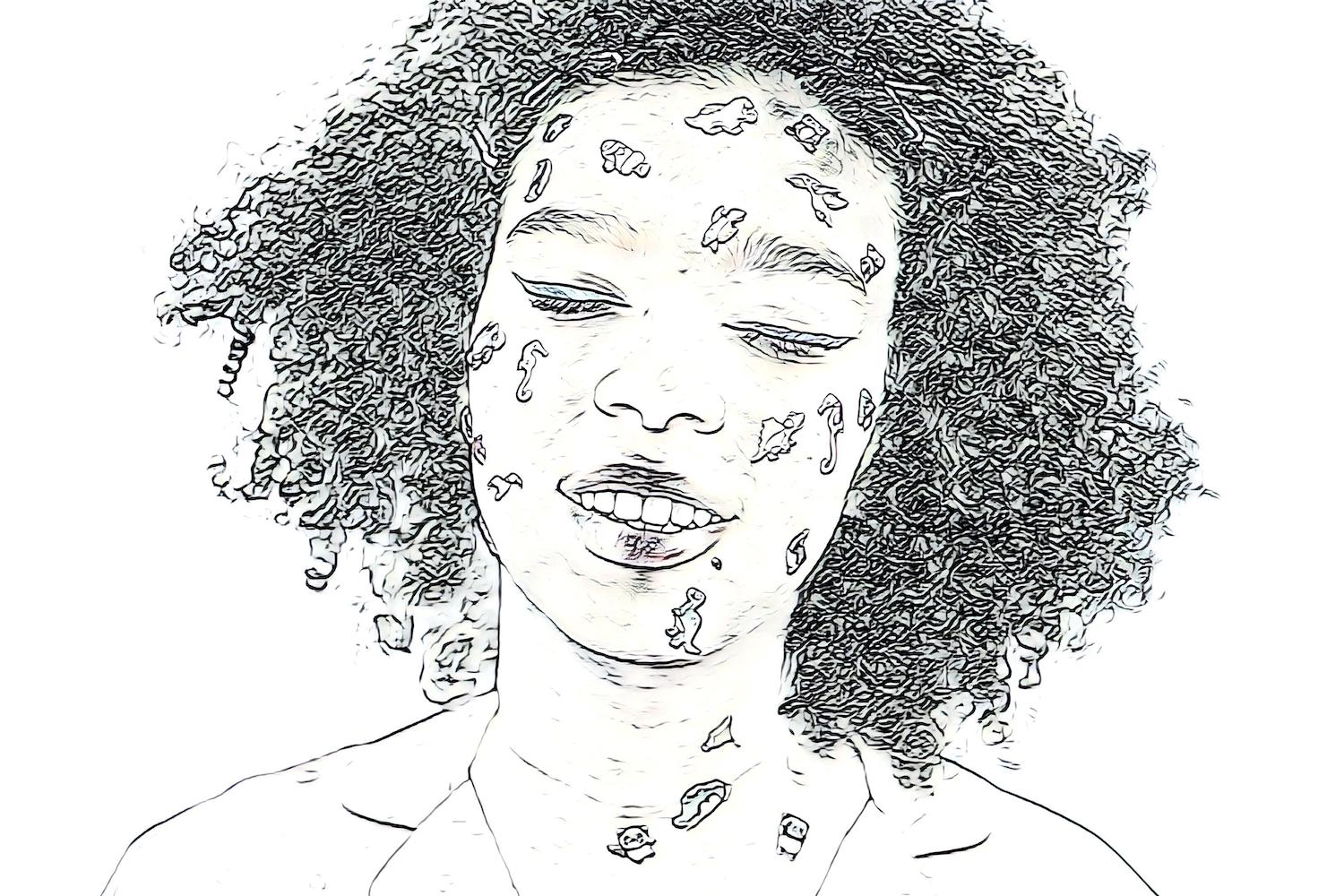}}&
		\multirow{2}{*}[0.386in]{\includegraphics[width=0.171\linewidth]{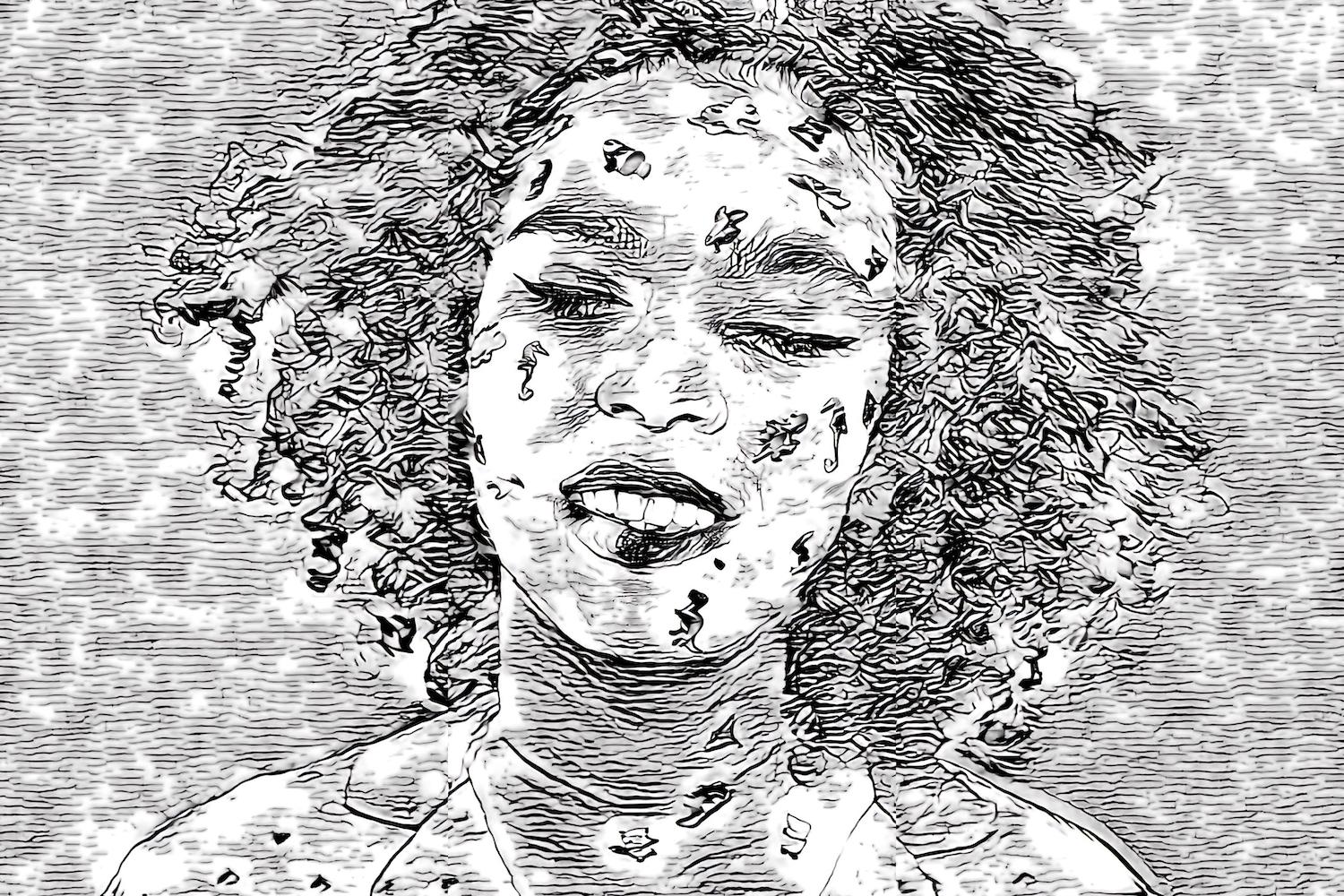}}&
		\multirow{2}{*}[0.386in]{\includegraphics[width=0.171\linewidth]{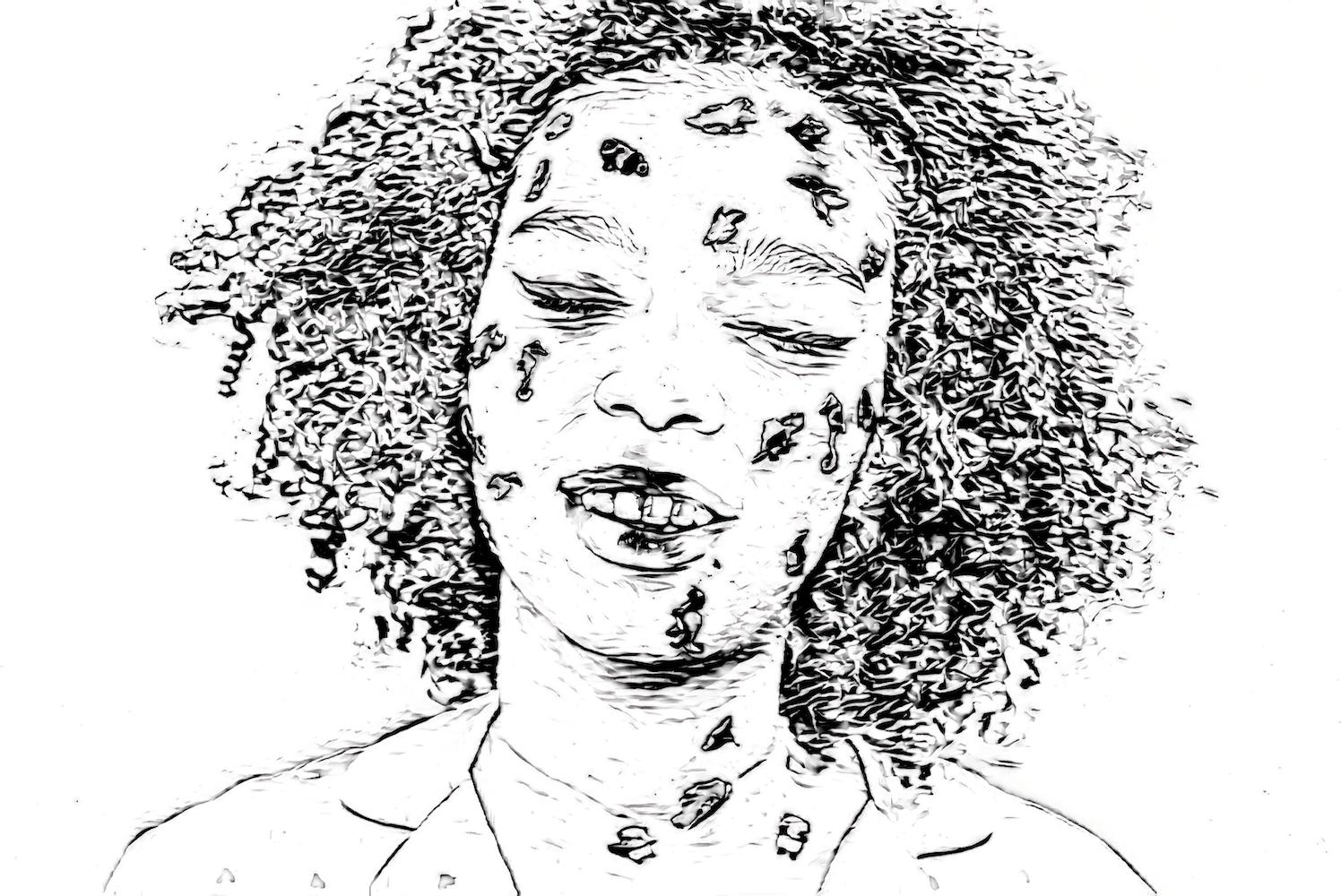}}&
		\multirow{2}{*}[0.386in]{\includegraphics[width=0.171\linewidth]{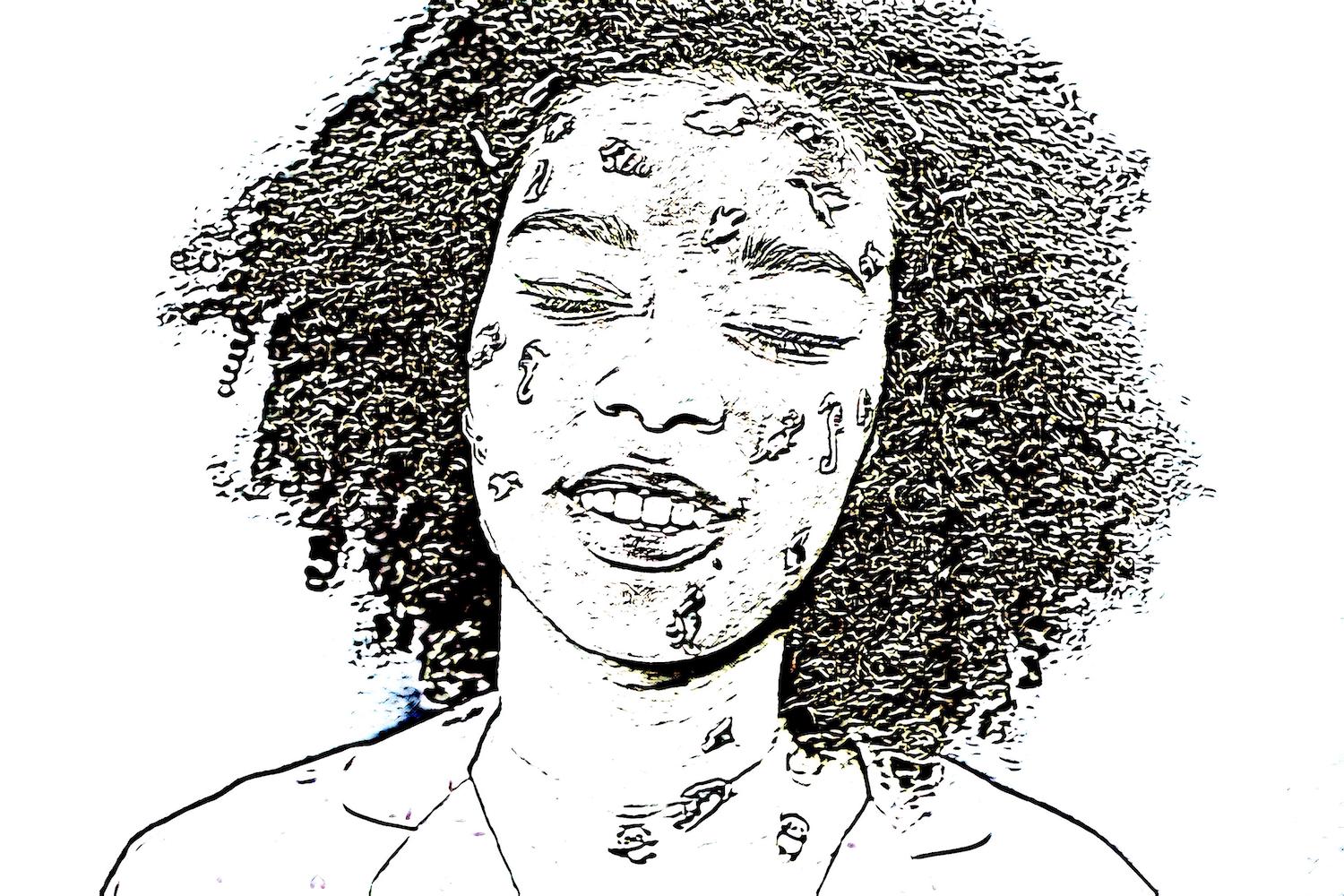}}
		\\
		\includegraphics[width=0.0851\linewidth]{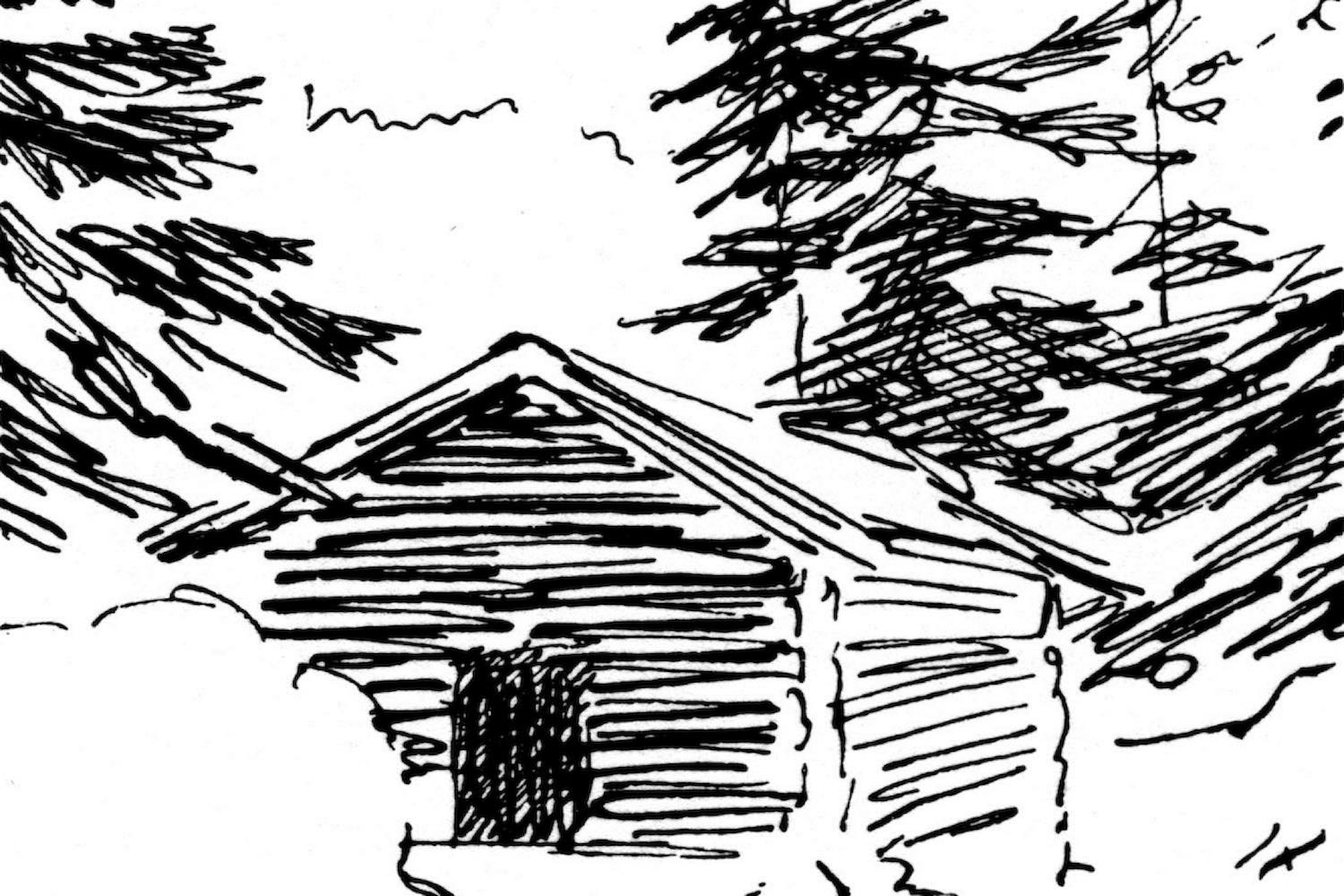}
		\\
		\\
		
		\includegraphics[width=0.0851\linewidth]{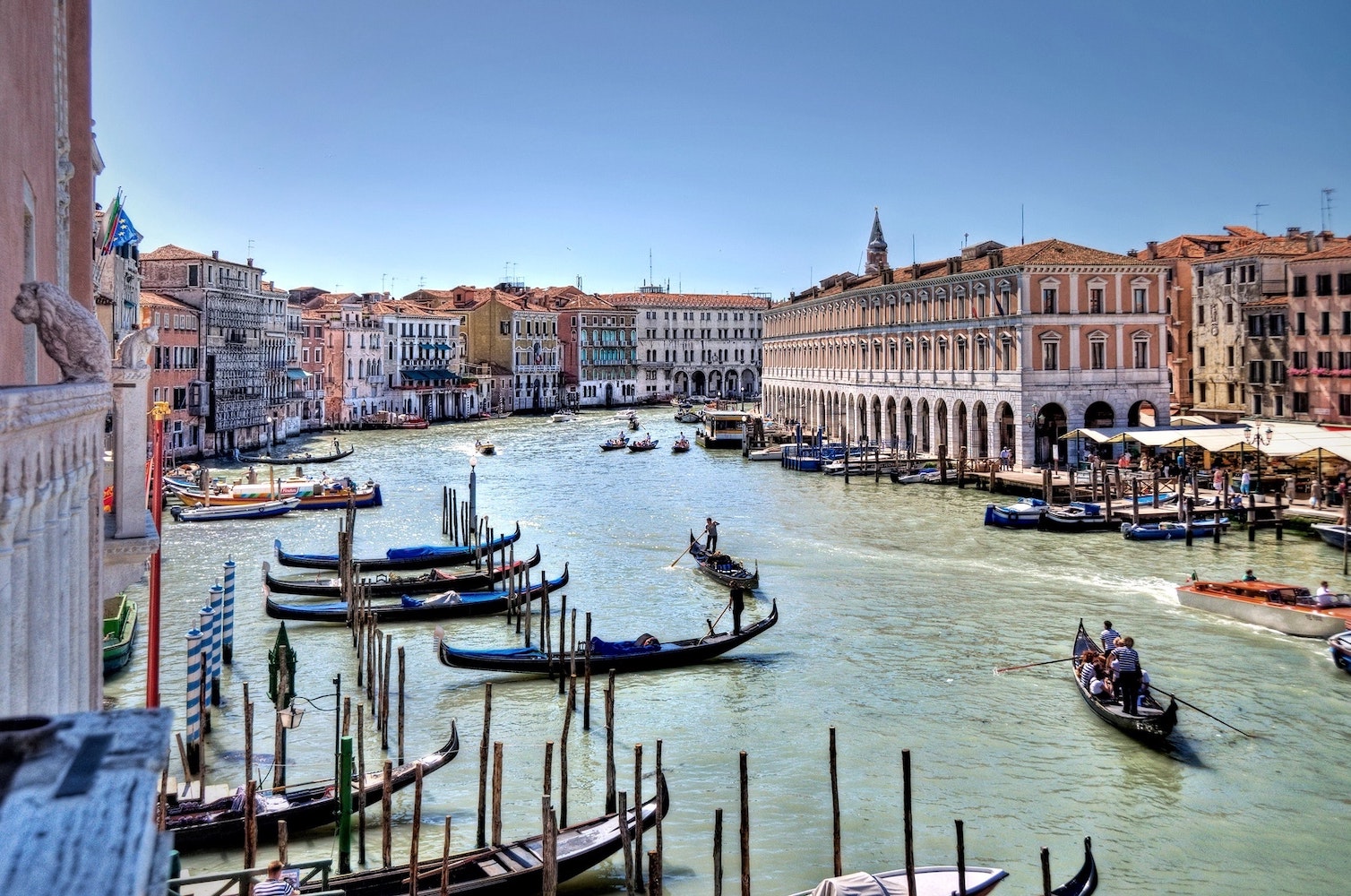}&
		\multirow{2}{*}[0.386in]{\includegraphics[width=0.171\linewidth]{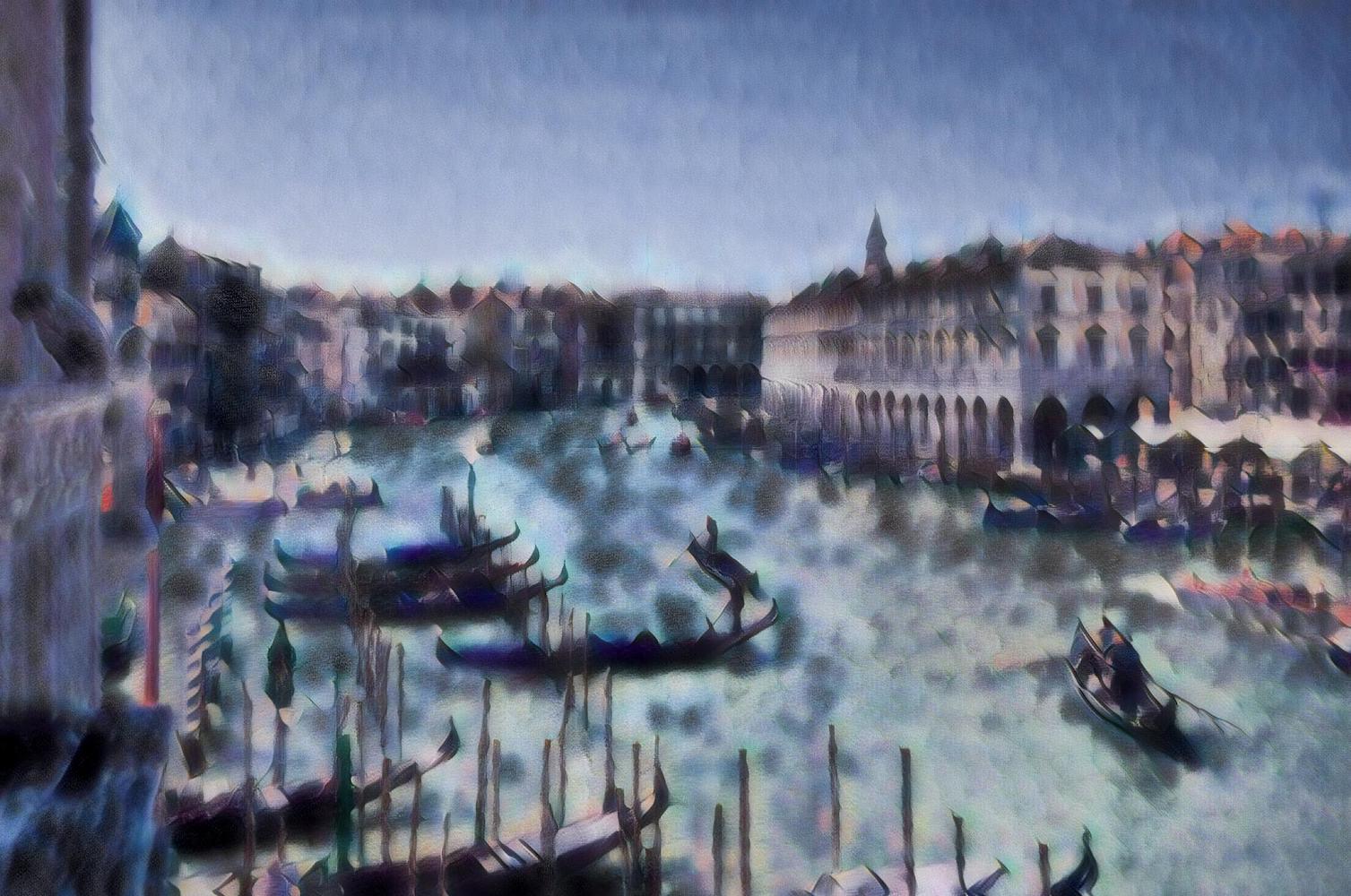}}&
		\multirow{2}{*}[0.386in]{\includegraphics[width=0.171\linewidth]{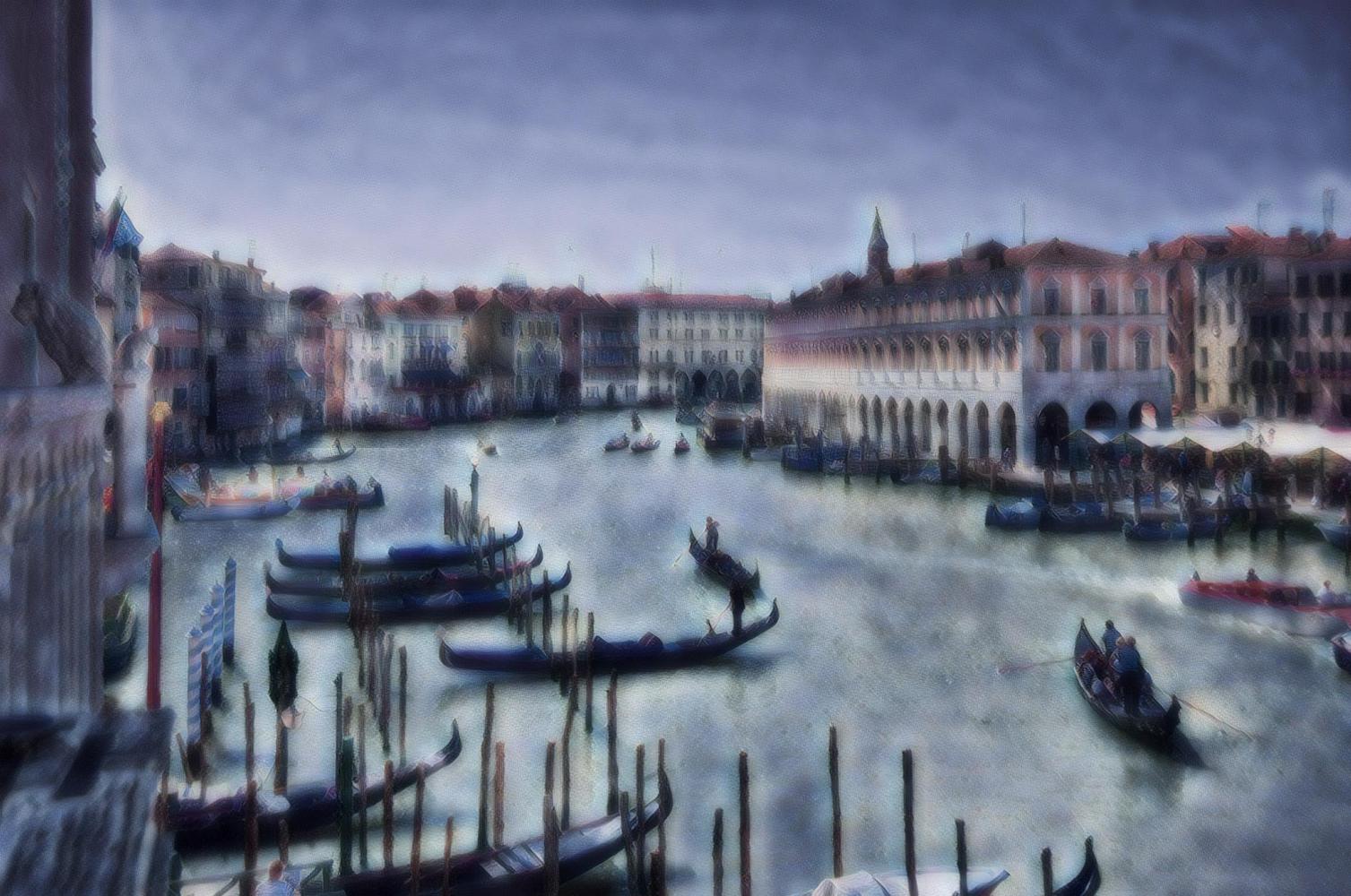}}&
		\multirow{2}{*}[0.386in]{\includegraphics[width=0.171\linewidth]{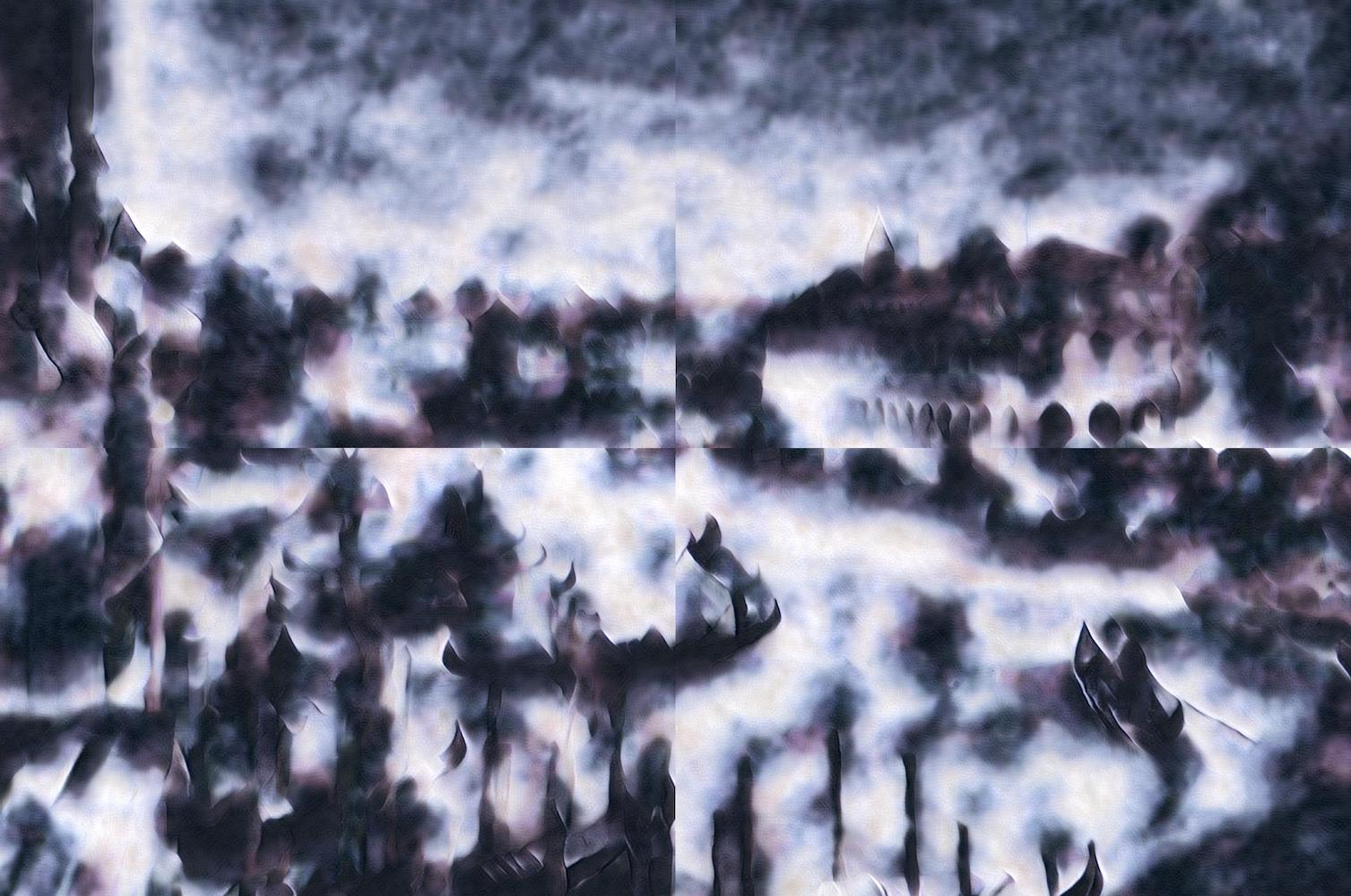}}&
		\multirow{2}{*}[0.386in]{\includegraphics[width=0.171\linewidth]{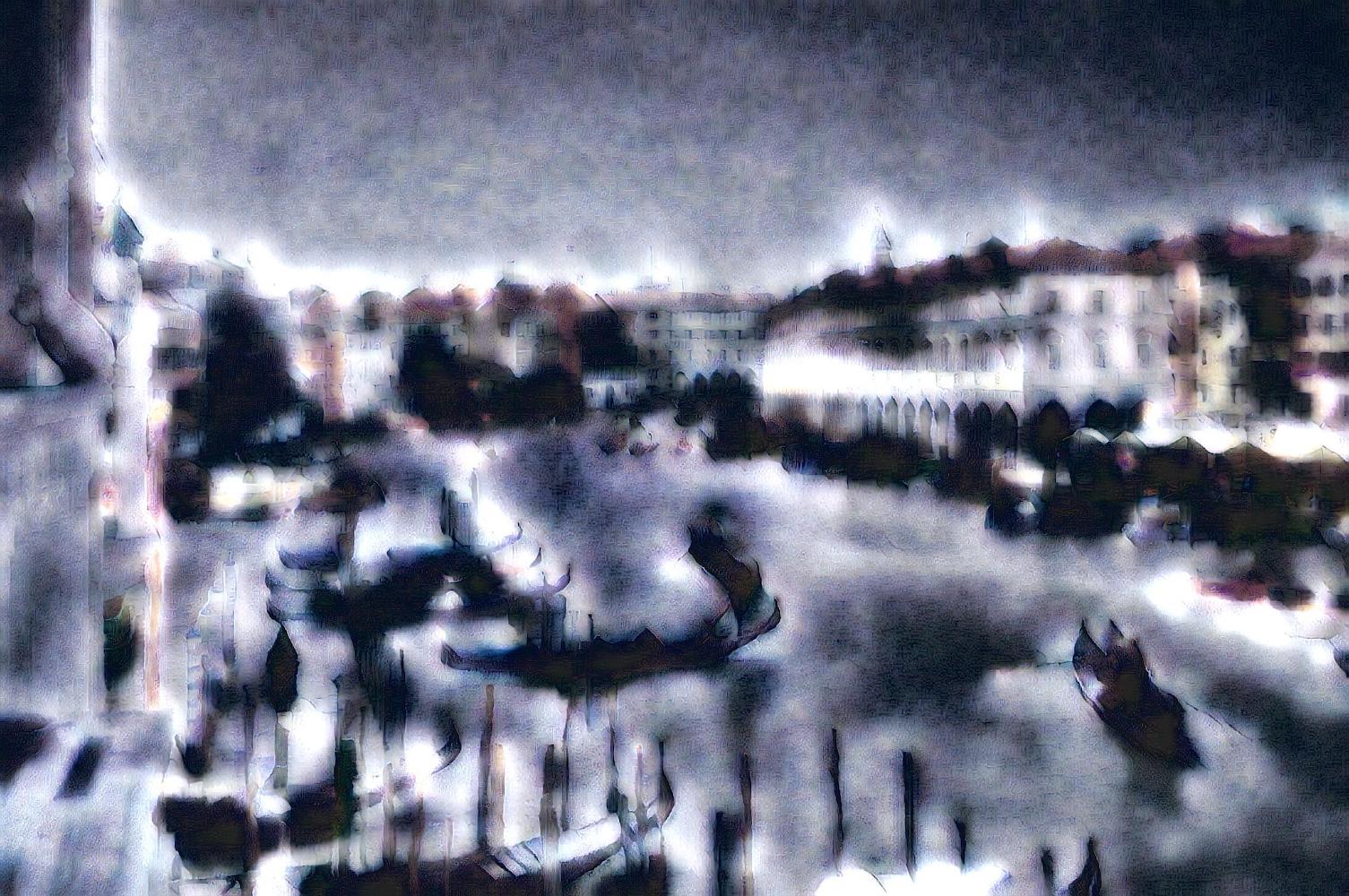}}&
		\multirow{2}{*}[0.386in]{\includegraphics[width=0.171\linewidth]{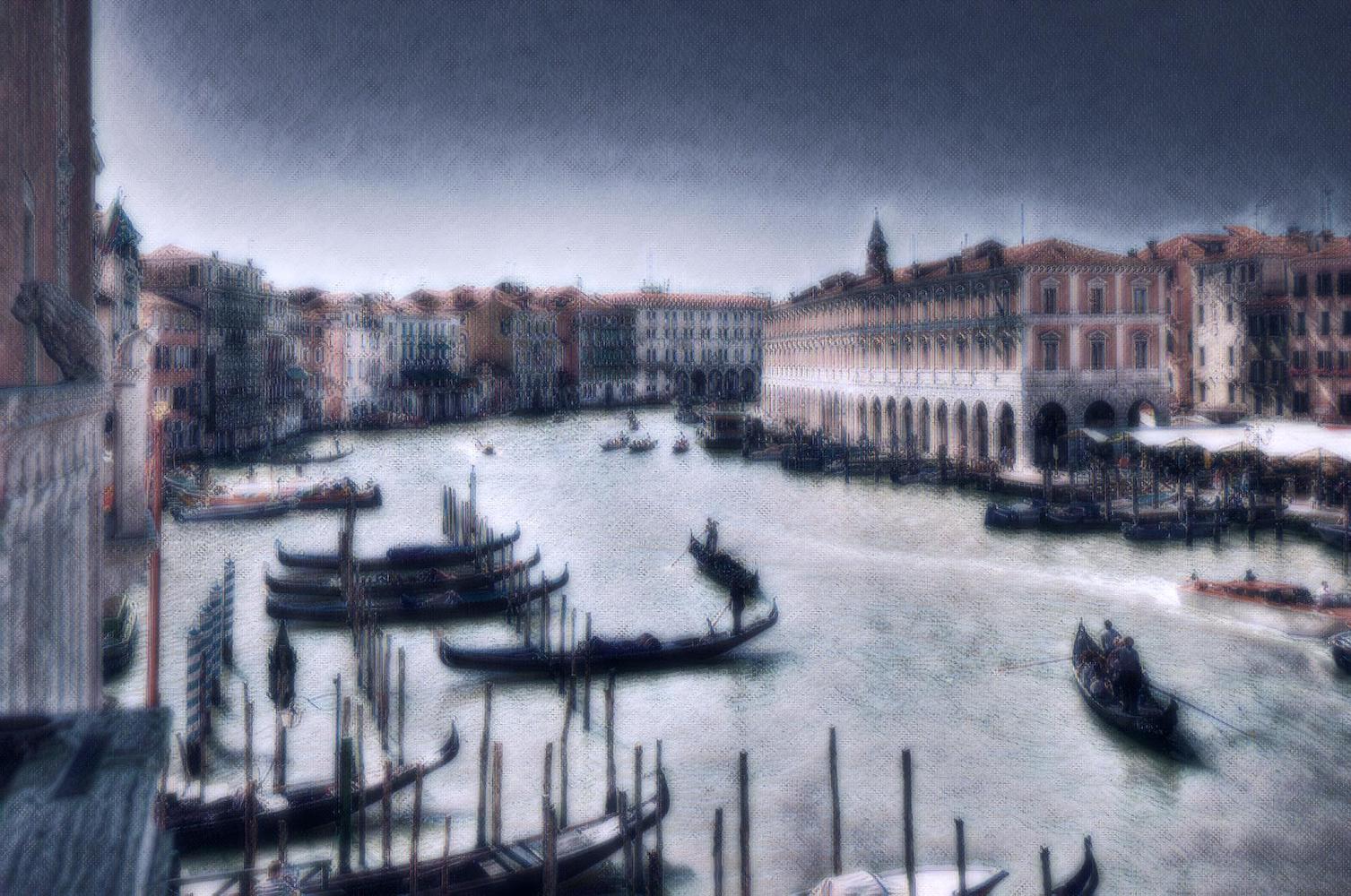}}
		\\
		\includegraphics[width=0.0851\linewidth]{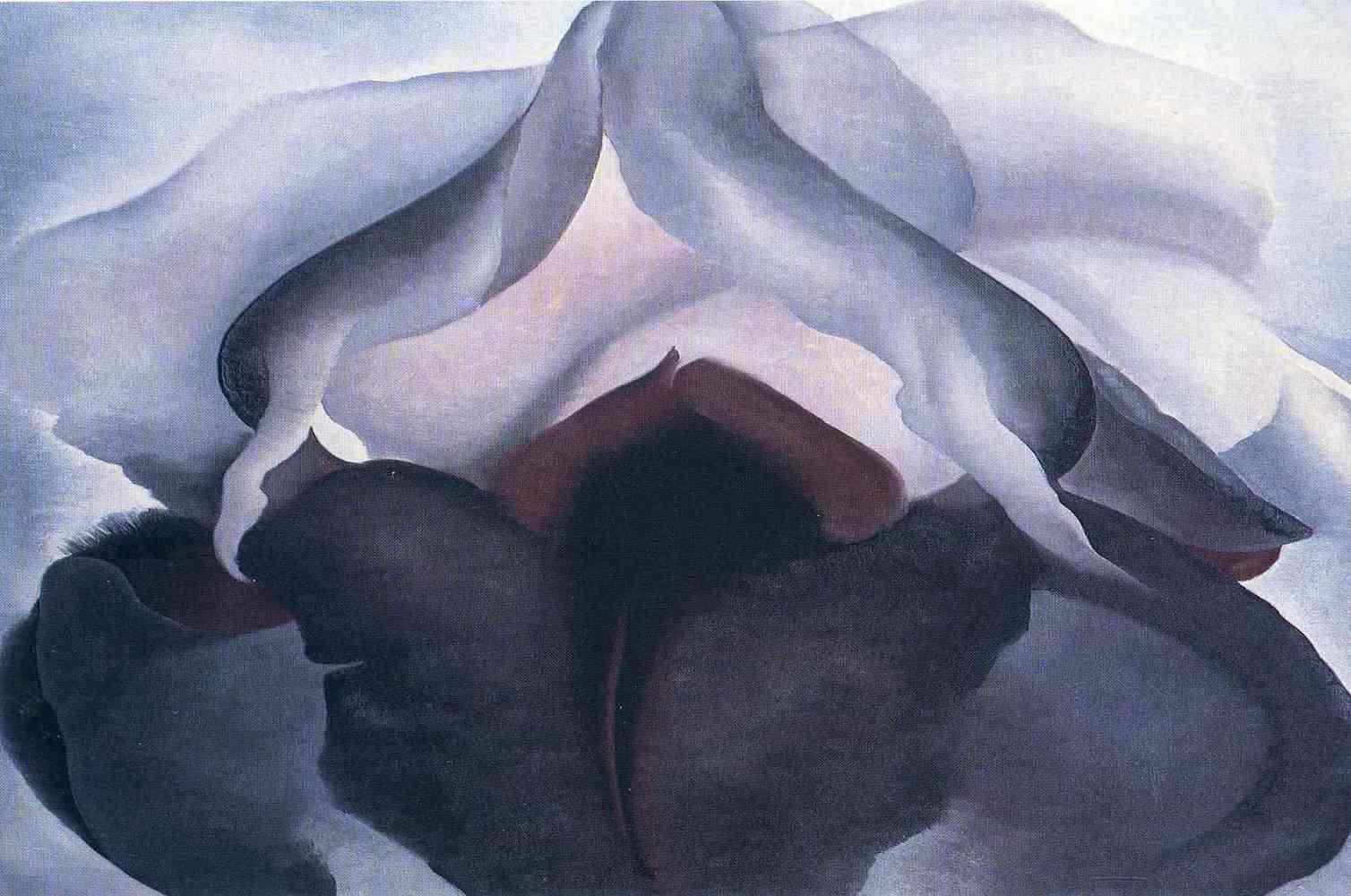}
		\\
		\\

		\includegraphics[width=0.0851\linewidth]{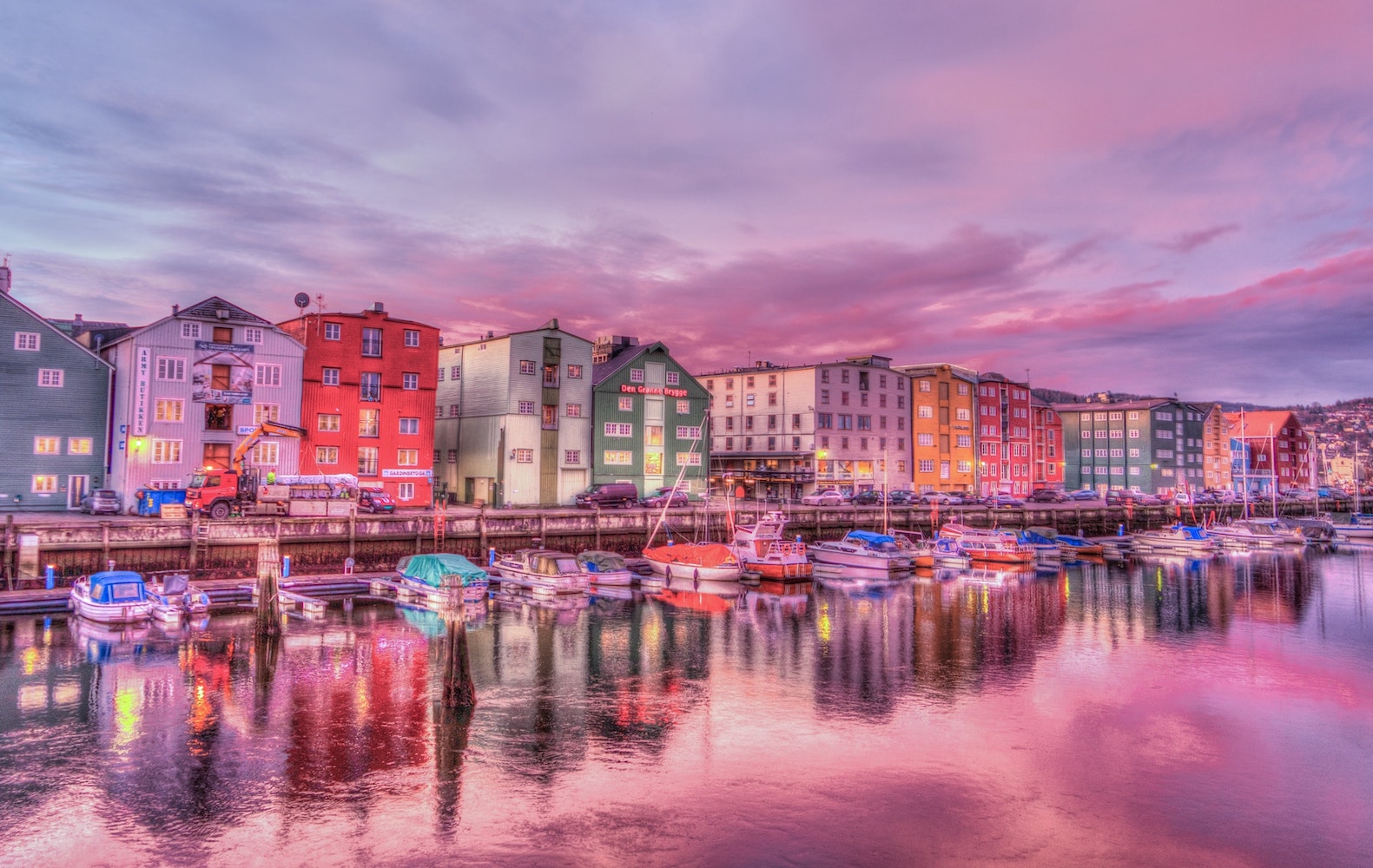}&
		\multirow{2}{*}[0.366in]{\includegraphics[width=0.171\linewidth]{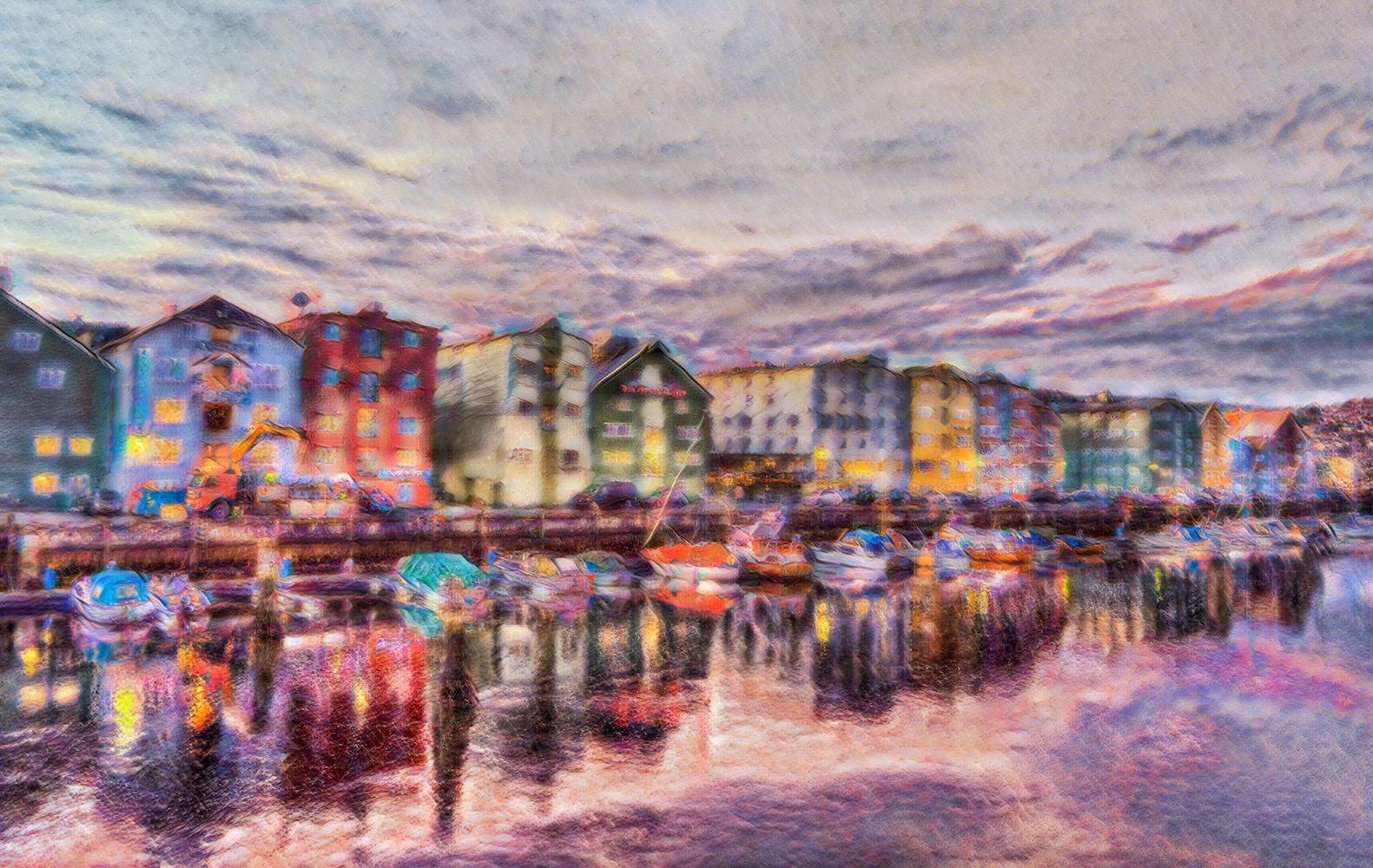}}&
		\multirow{2}{*}[0.366in]{\includegraphics[width=0.171\linewidth]{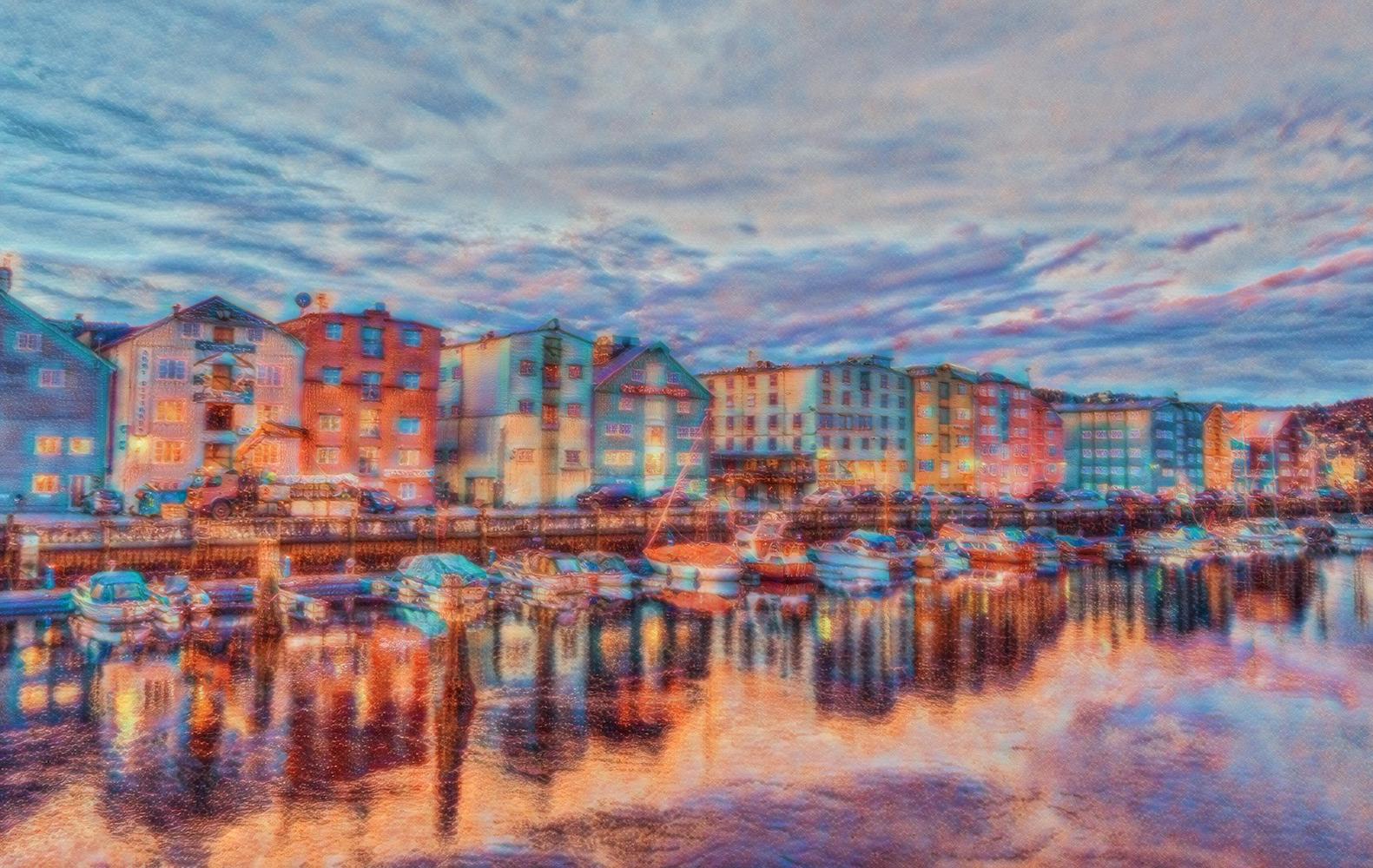}}&
		\multirow{2}{*}[0.366in]{\includegraphics[width=0.171\linewidth]{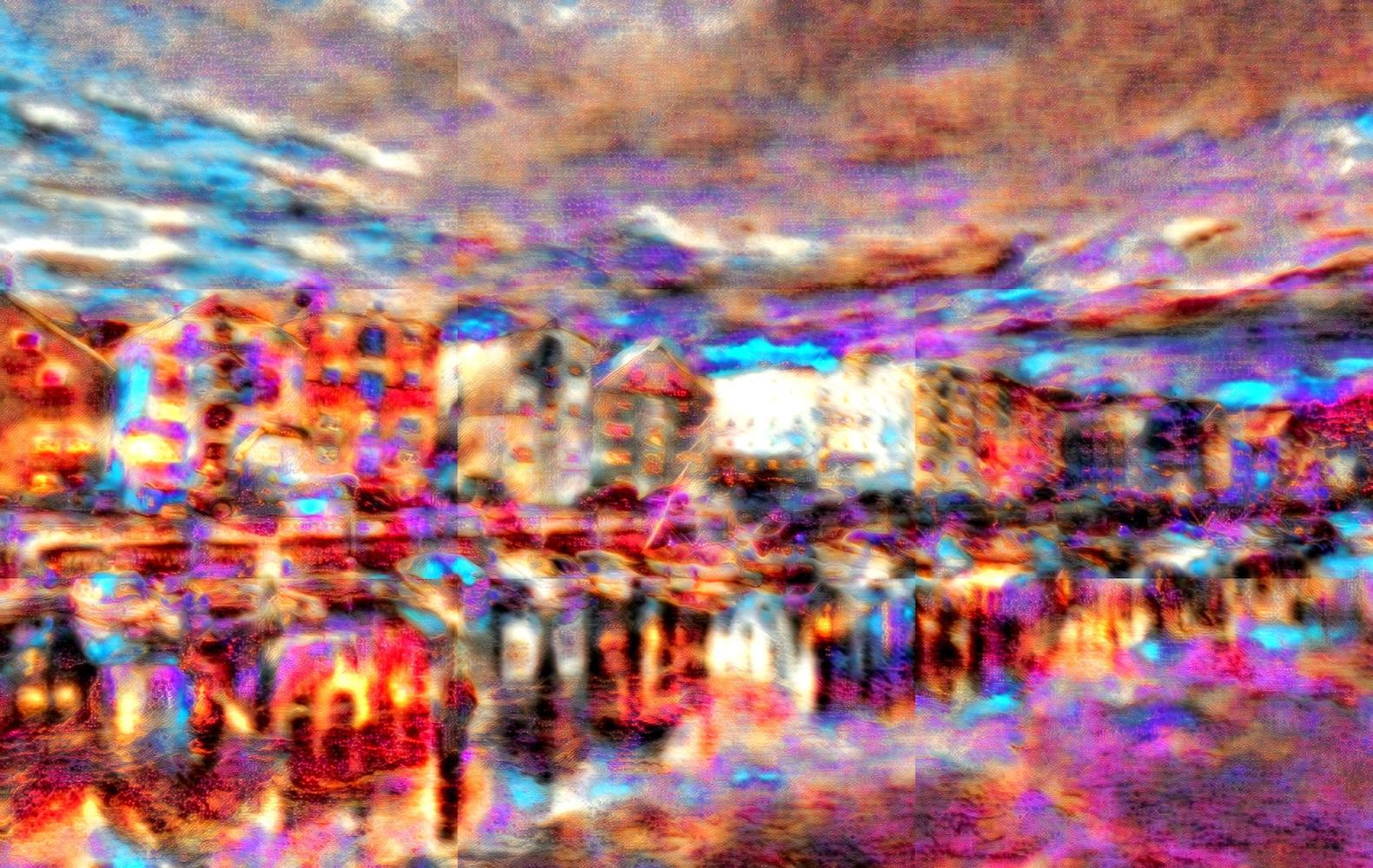}}&
		\multirow{2}{*}[0.366in]{\includegraphics[width=0.171\linewidth]{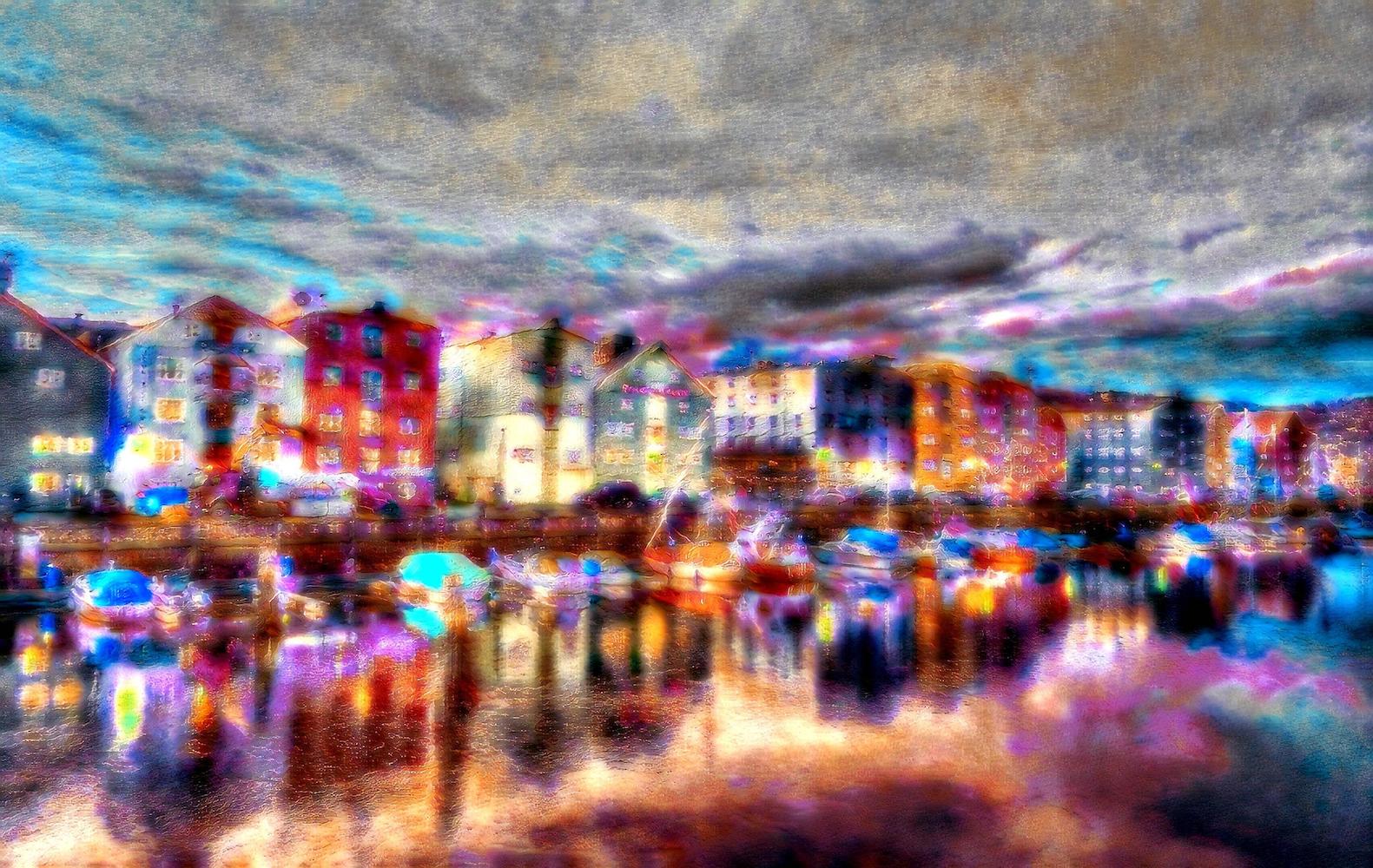}}&
		\multirow{2}{*}[0.366in]{\includegraphics[width=0.171\linewidth]{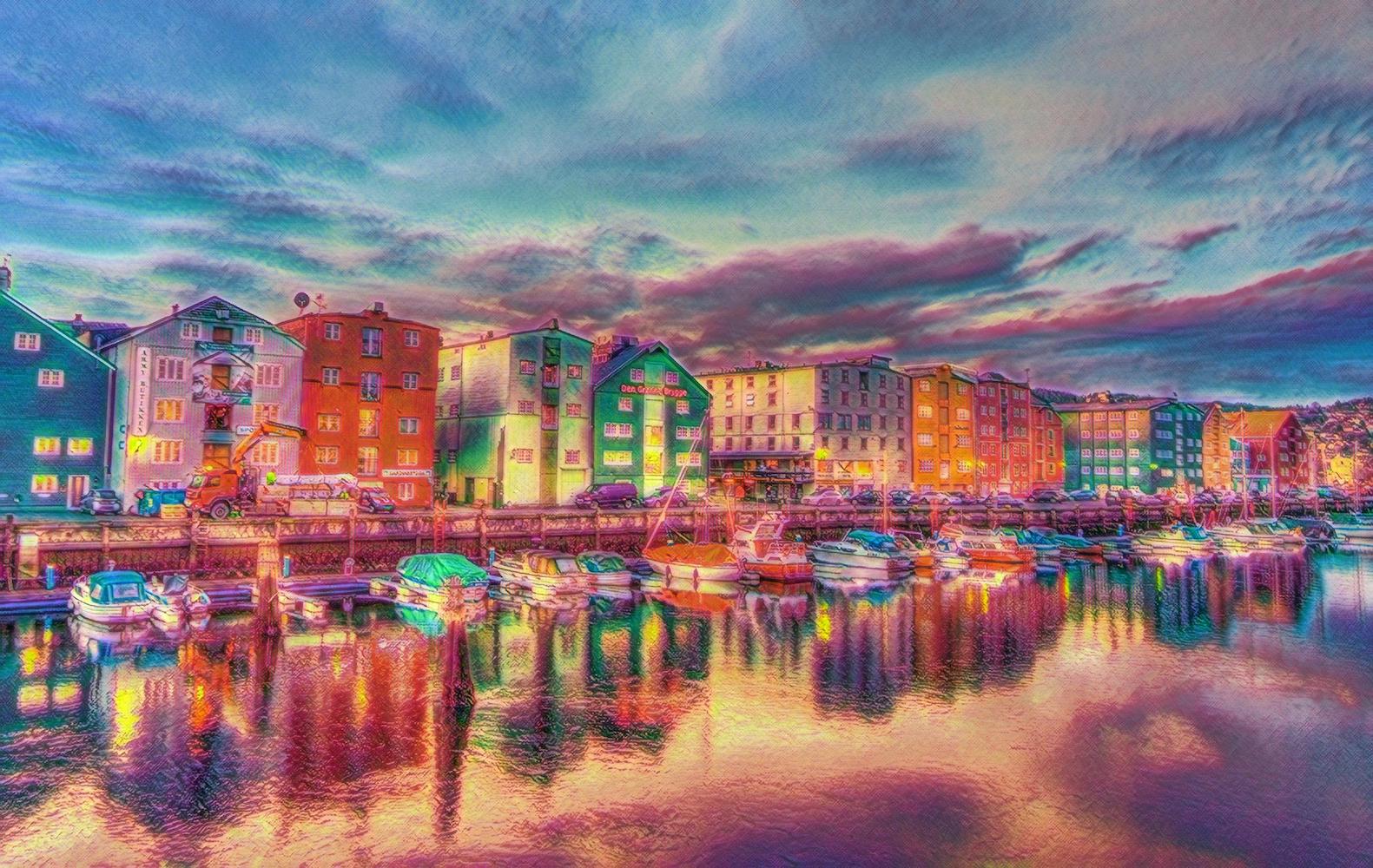}}
		\\
		\includegraphics[width=0.0851\linewidth]{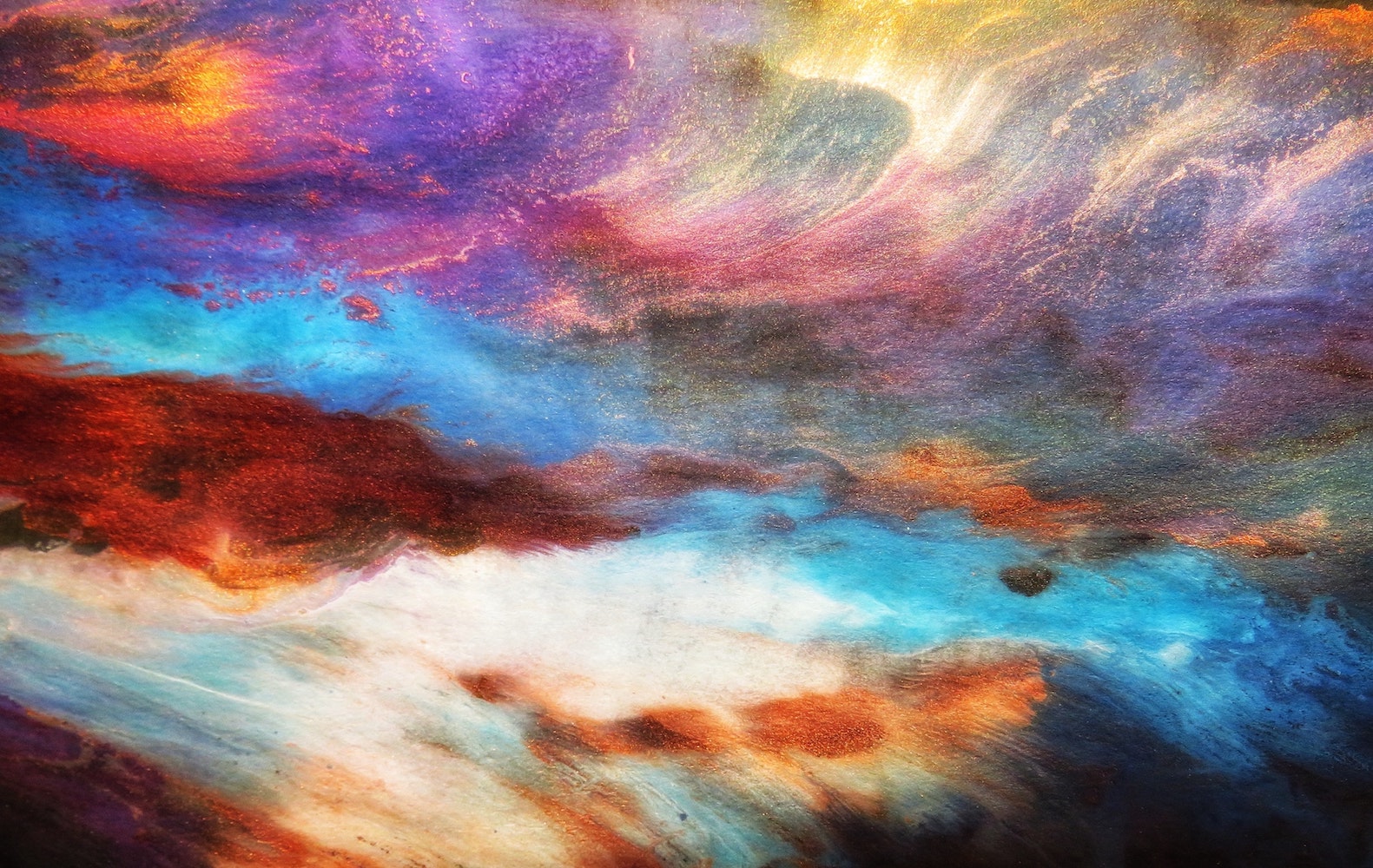}
		\\
		\\
		
		\includegraphics[width=0.0851\linewidth]{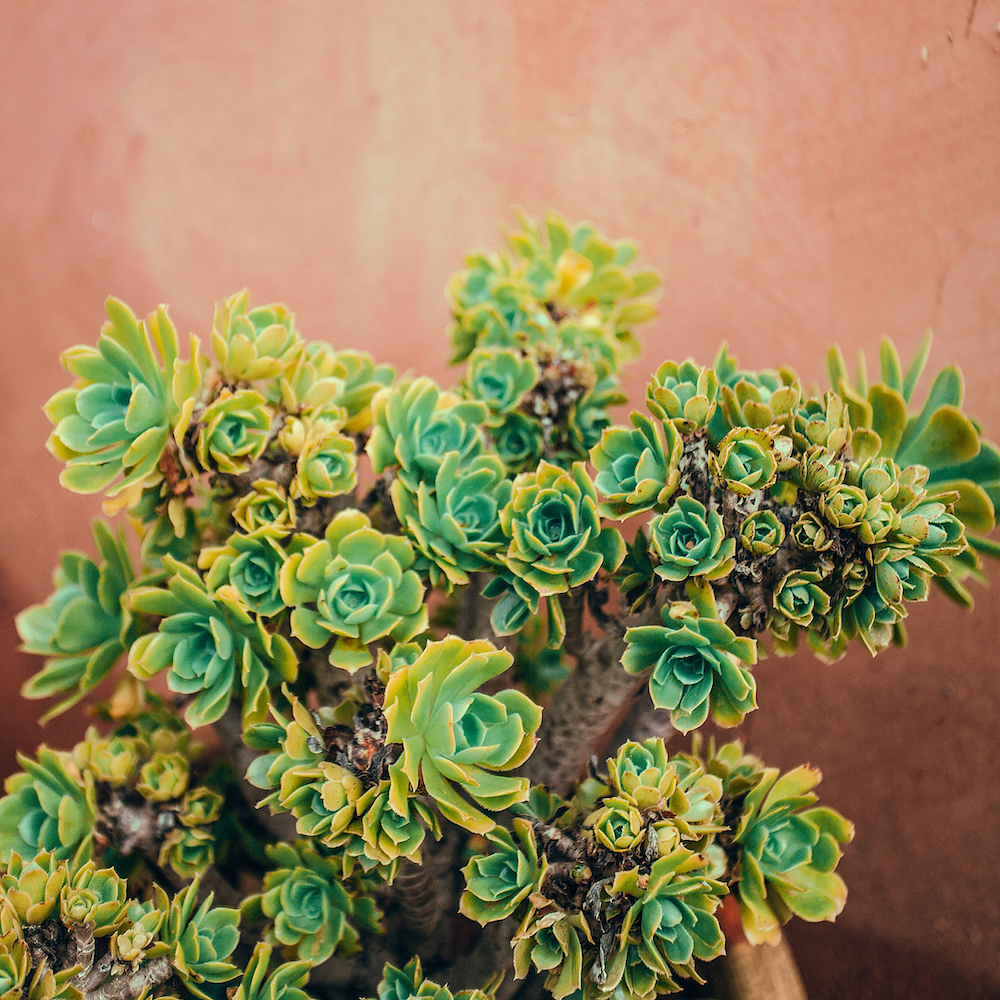}&
		\multirow{2}{*}[0.584in]{\includegraphics[width=0.171\linewidth]{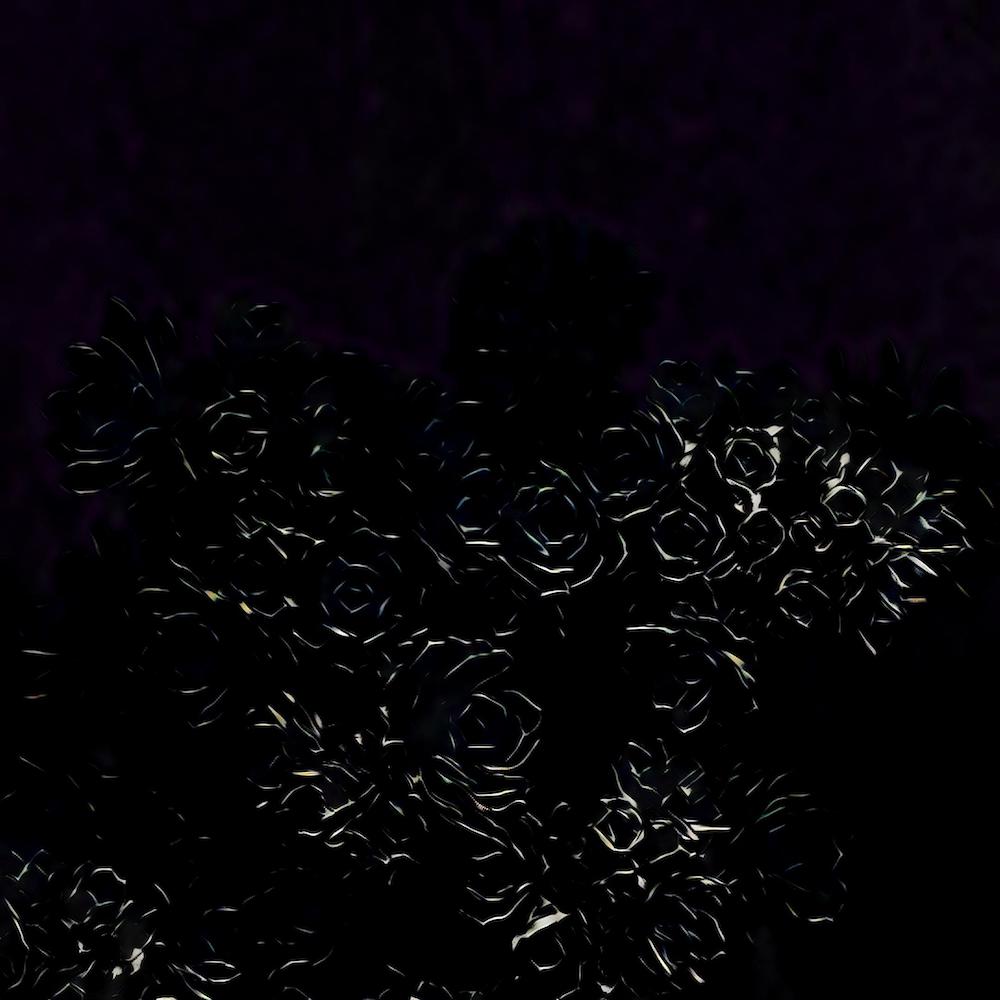}}&
		\multirow{2}{*}[0.584in]{\includegraphics[width=0.171\linewidth]{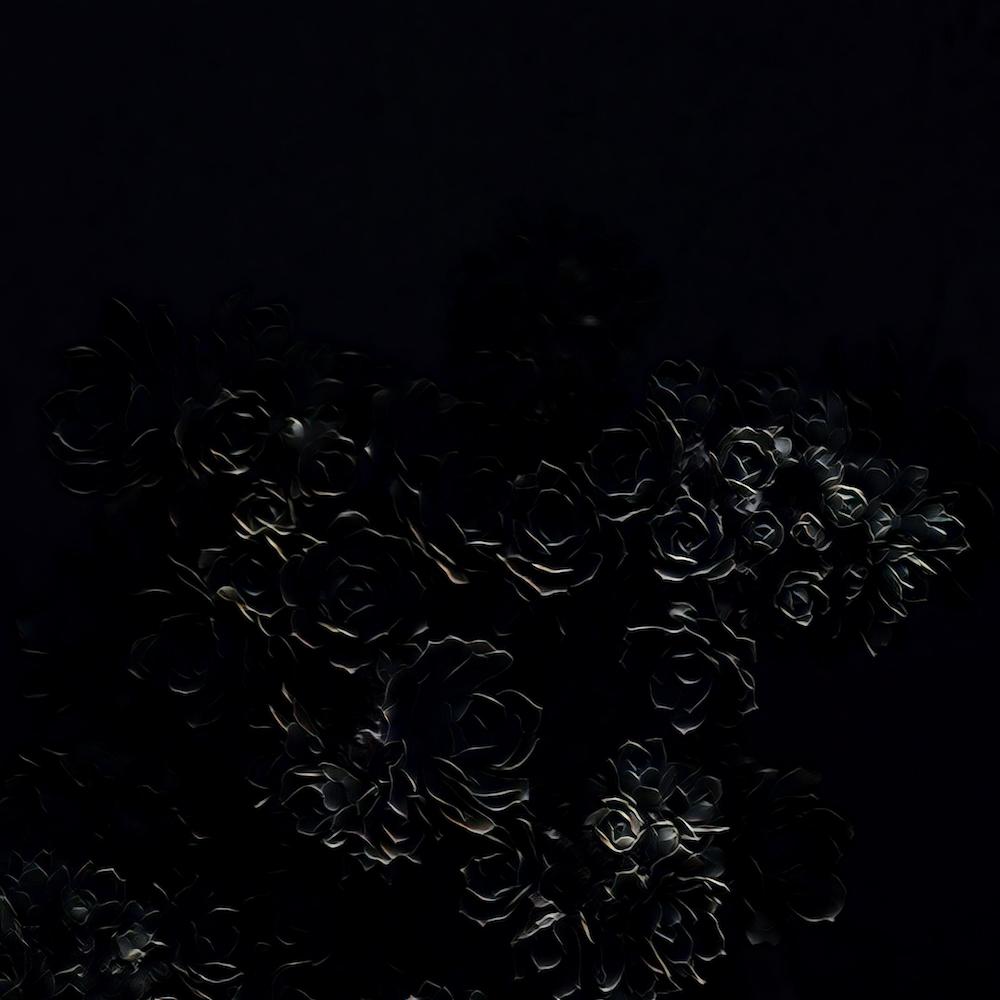}}&
		\multirow{2}{*}[0.584in]{\includegraphics[width=0.171\linewidth]{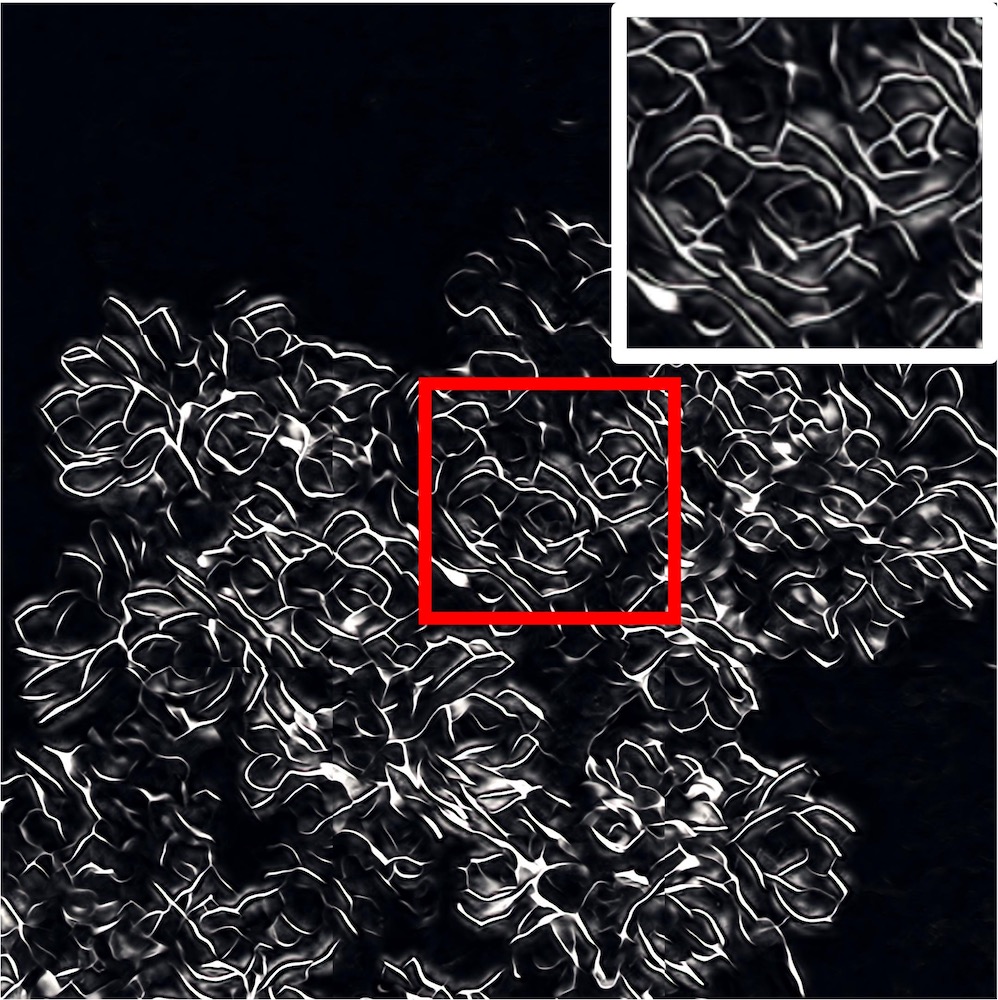}}&
		\multirow{2}{*}[0.584in]{\includegraphics[width=0.171\linewidth]{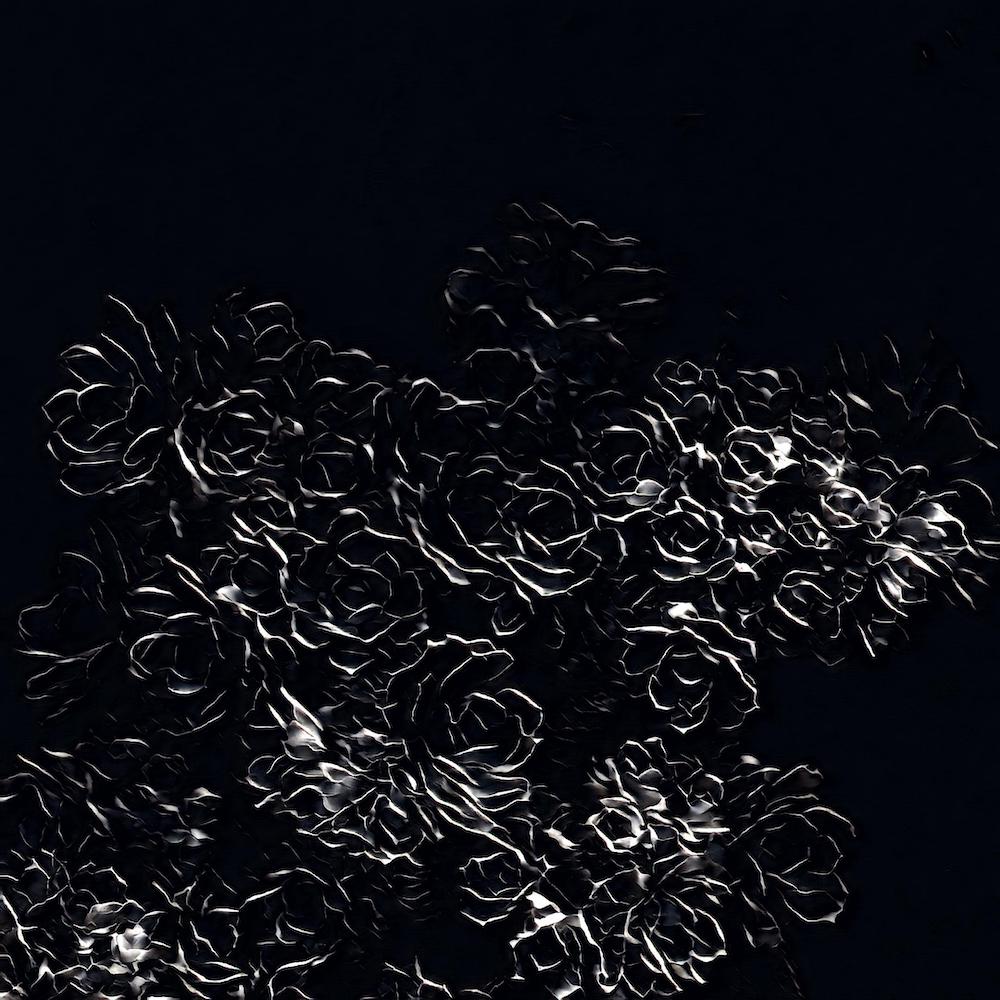}}&
		\multirow{2}{*}[0.584in]{\includegraphics[width=0.171\linewidth]{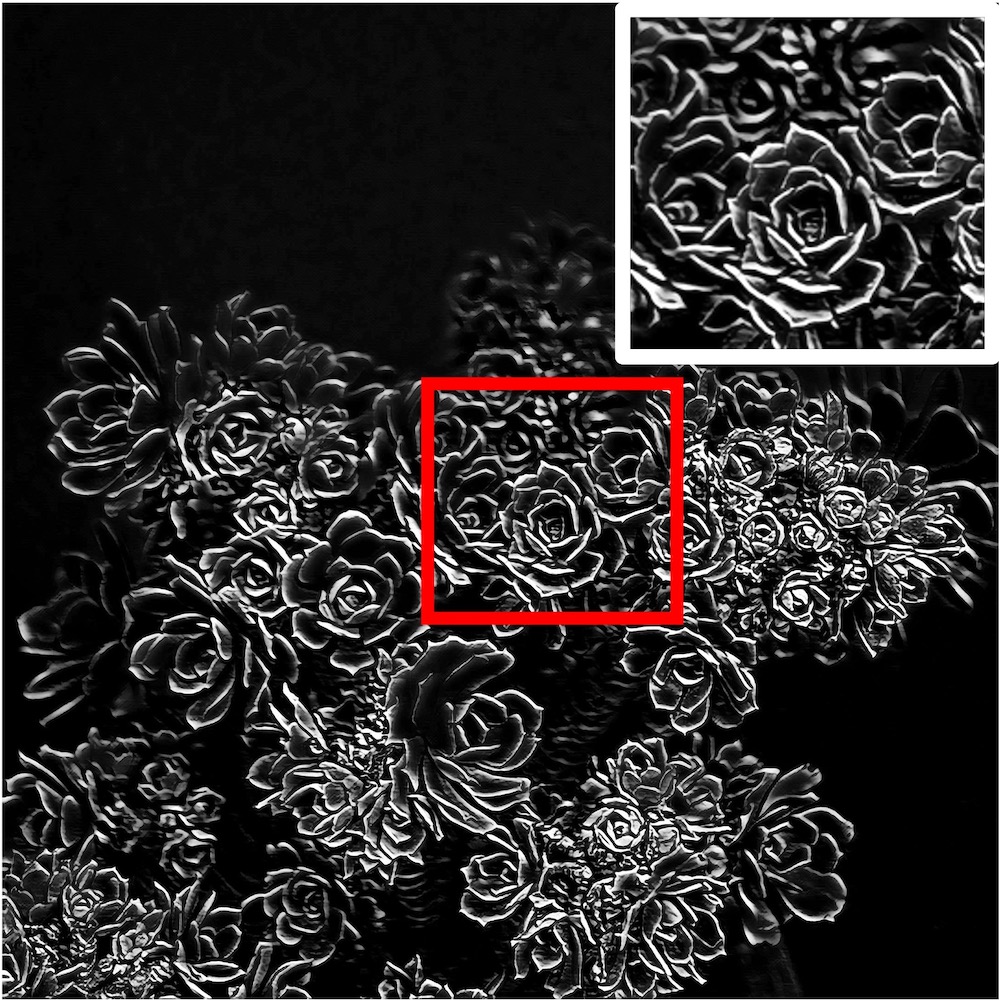}}
		\\
		\includegraphics[width=0.0851\linewidth]{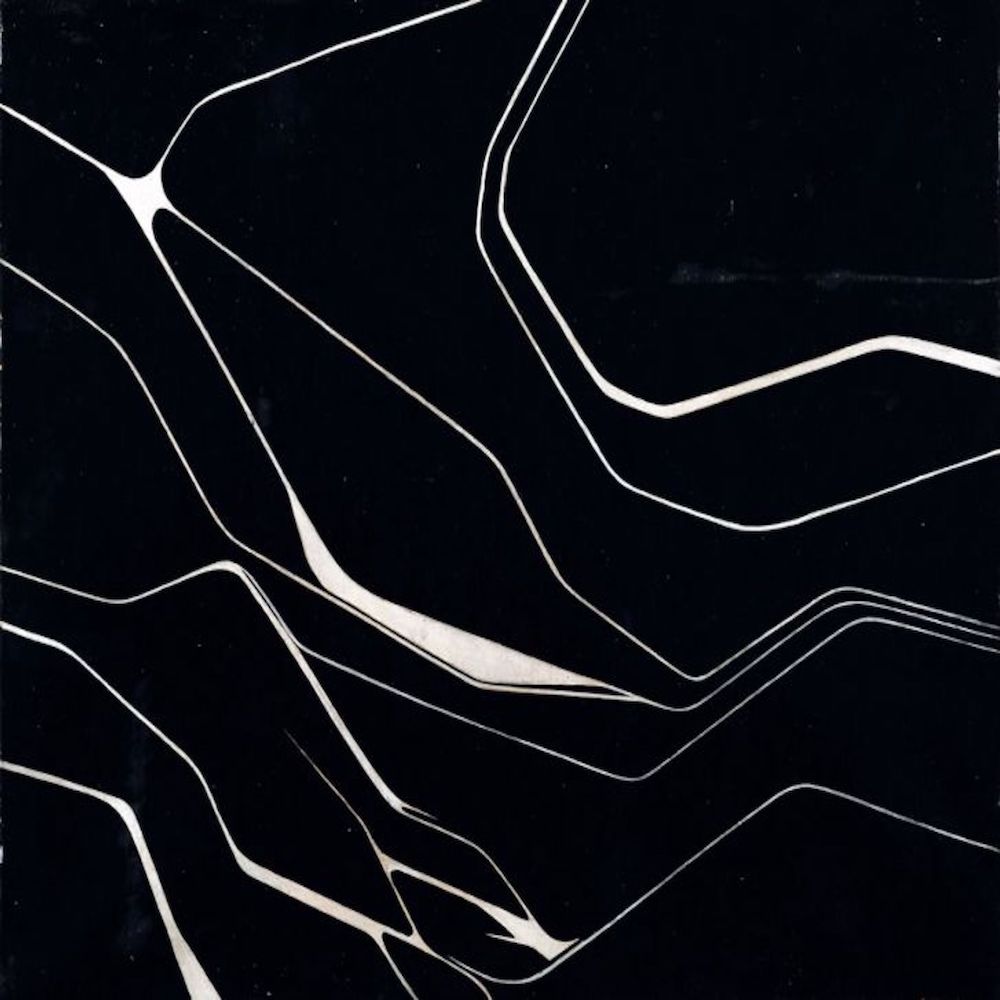}
		\\
		\\
		\\
		\footnotesize Inputs & \footnotesize AdaIN-U & \footnotesize LST-U & \footnotesize WCT-U & \footnotesize  Collab-Distill  &  \footnotesize {\bf Our MicroAST}
		
	\end{tabular}
	\caption{{\bf Qualitative comparison} with the state of the art on ultra-resolution (4K) images. See more in {\em SM}.
	}
	\label{fig:cmp}
\end{figure*}

\renewcommand\arraystretch{0.8}
\begin{table*}[t]
	\centering
	\setlength{\tabcolsep}{0.19cm}
	\begin{tabular}{c|ccccccc}
		\toprule
		\footnotesize Method & \footnotesize (a) \#Params/$10^6$  & \footnotesize (b) Storage/MB & \footnotesize (c) GFLOPs & \footnotesize  (d) Time/sec  &  \footnotesize (e) SSIM $\uparrow$ & \footnotesize  (f) Style Loss $\downarrow$ &  \footnotesize  (g) Preference/\%
		\\
		\midrule
		\footnotesize AdaIN-U& \footnotesize 7.011 & \footnotesize 94.100 & \footnotesize 5841.9 & \footnotesize 2.884  &  \footnotesize  0.322 & \footnotesize  2.401 &  \footnotesize 10.0
		\\
		\footnotesize LST-U & \footnotesize 12.167 & \footnotesize 48.600 & \footnotesize  6152.1 & \footnotesize 2.892  &  \footnotesize 0.441 & \footnotesize 2.420 &  \footnotesize 24.2
		\\
		\footnotesize WCT-U & \footnotesize 34.239 & \footnotesize 120.268 & \footnotesize 19390.1 & \footnotesize 9.528 &  \footnotesize 0.248  &  \footnotesize 2.226 &  \footnotesize 16.2
		\\
		\footnotesize Collab-Distill & \footnotesize 2.146 & \footnotesize 9.659 & \footnotesize 1338.9 & \footnotesize 7.139 &  \footnotesize 0.292 &  \footnotesize \bf 2.183  &  \footnotesize 20.5
		\\
		\midrule
		\footnotesize {\bf MicroAST} & \footnotesize \bf 0.472 & \footnotesize \bf 1.857 & \footnotesize \bf 374.9 & \footnotesize \bf 0.522  & \footnotesize \bf 0.531  &  \footnotesize 2.342  &  \footnotesize \bf 29.1
		
		\\
		\bottomrule
		
	\end{tabular}
	\caption{{\bf Quantitative comparison} with the state of the art. Storage is measured in PyTorch model. GFLOPs and Time are measured when the content and style are both 4K images and tested on an NVIDIA RTX 2080 (8GB) GPU. The best results are set in {\bf bold}. $\uparrow$: Higher is better. $\downarrow$: Lower is better.}
	\label{tab:quantity}
\end{table*}

\subsection{Style Signal Contrastive Learning}
\label{SSCL}
As analyzed in Sec.~\ref{DM}, due to the simplicity of network architecture, the micro style encoder $E_s$ has limited ability to extract sufficiently complex style representations. Inspired by recent contrastive learning~\cite{he2020momentum,chen2020simple,tian2020contrastive,wu2021contrastive} which aims at improving the representative power of neural networks, we propose a novel {\em style signal contrastive (SSC) loss} $\mathcal{L}_{ssc}$ to boost the style representative ability of $E_s$.

The core idea of contrastive learning is to pull data points (called ``query'') close to their ``positive” examples, while pushing them apart from other examples that are regarded as ``negatives” in the representation space. Therefore, how to construct the ``positive” pairs and ``negative” pairs is a key problem we need to consider. Intuitively, in our MicroAST, every stylized result is generated under the modulations of a set of specific style signals extracted from a target style image. Therefore, it should possess the similar style signals with the target style image while exhibiting the distinct style signals from other style images. Based on this intuition, given a training mini-batch including $N$ content images $\phi_c = \{C_1, C_2, \dots, C_N\}$ and $N$ style images $\phi_s =\{S_1, S_2, \dots, S_N\}$, we first generate $N$ stylized outputs $ \phi_o = \{O_1, O_2, \dots, O_N\}$ ($O_i$ is generated by using $C_i$ as content and $S_i$ as style). Then, for each ``query'' $O_i$, we can construct $\phi_p = \{S_i\}$ as its ``positive” example, and $\phi_n = \{S_j \in \phi_s | j\neq i\}$ as ``negative” examples. Finally, our SSC loss $\mathcal{L}_{ssc}$ is formulated based on the extracted style modulation signals (Eq.~(\ref{eq:dm})) of them.
\begin{equation}
	\mathcal{L}_{ssc} := \sum_{i=1}^{N} \frac{\parallel m_{o_i} -m_{s_i}\parallel_2 }{\sum_{j\neq i}^{N} \parallel m_{o_i} -m_{s_j}\parallel_2}.
\end{equation}

$\mathcal{L}_{ssc}$ plays a role of {\em opposing forces} pulling the style signals of $O_i$ to those of its target style image $S_i$, and pushing them away from those of other style images. Therefore, it could boost the ability of the micro style encoder to extract more distinct and representative style modulation signals, further improving the stylization quality (see Sec.~\ref{ablation}).

{\bf Discussion.} Recently, \cite{chen2021artistic} also introduced contrastive learning for style transfer. There are, however, three main differences: (1) Their contrastive losses are optimized on the {\em generator} of SANet~\cite{park2019arbitrary} to improve the quality, and their encoders are fixed pre-trained VGG. In contrast, our $\mathcal{L}_{ssc}$ is optimized mainly on the {\em micro style encoder} to boost its ability in ultra-resolution style transfer. (2) They construct the ``positive” pairs and ``negative” pairs only within the stylized results, while ours are constructed between the style images and the stylized results. (3) Their contrastive losses are vanilla InfoNCE losses~\cite{oord2018representation}, which we found is less effective in our task (almost no improvement). Thus, we provide a different form of contrastive loss for our $\mathcal{L}_{ssc}$, which is more straightforward and effective in our task.

\cite{zhang2022domain} and \cite{wu2022ccpl} are two concurrent works using contrastive learning in style transfer. Apart from different goals of contrastive learning, they still rely on large pre-trained DCNNs as encoders, and their contrastive losses remain vanilla InfoNCE losses.

%\vspace{-0.7em}
\section{Experimental Results}
\label{exp}

%\vspace{-0.1em}
\subsection{Implementation Details}
\label{impd}
We implement a multi-level DualMod which modulates both the two ResBlocks of the micro decoder $D$ (we omit the modulations for ResBlock\_2 in Fig.~\ref{fig:overview} and Fig.~\ref{fig:mod} for brevity). The loss weights in Eq.~(\ref{eq:loss}) are set to $\lambda_c=1$, $\lambda_s=3$, and $\lambda_{ssc}=3$. We train our MicroAST using MS-COCO~\cite{lin2014microsoft} as content images and WikiArt~\cite{phillips2011wiki} as style images. We use the Adam optimizer~\cite{kingma2014adam} with a learning rate of 0.0001 and a mini-batch size of 8 content-style image pairs. During training, all images are loaded with the smaller dimension rescaled to 512 pixels while preserving the aspect ratio, and then randomly cropped to 256$\times$256 pixels for augmentation. Since our MicroAST is fully convolutional, it can handle any input size during testing. All experiments are conducted on an NVIDIA RTX 2080 (8GB) GPU.

%\vspace{-0.5em}
\subsection{Comparisons with Prior Arts}
Since our goal is to achieve ultra-resolution AST, we compare our MicroAST with two types of state-of-the-art ultra-resolution AST methods based on (1) model compression~\cite{wang2020collaborative} and (2) patch-wise stylization~\cite{chen2022towards}. We directly run the author-released codes with default settings for the compared methods.

{\bf Qualitative Comparison.} Fig.~\ref{fig:cmp} shows the qualitative comparison results. The patch-wise stylization of URST \cite{chen2022towards} (marked by ``-U'') can help existing AST methods AdaIN~\cite{huang2017arbitrary}, WCT~\cite{li2017universal}, and LST~\cite{li2019learning} achieve ultra-resolution style transfer. However, AdaIN-U and LST-U often produce less stylized results which retain the colors of content images ({\em e.g.}, the hair and eyelash colors in the $1^{st}$ row, and the river color in the $2^{nd}$ row) or transfer insufficient colors of style images ({\em e.g.}, the $3^{rd}$ row). WCT-U can transfer more faithful colors, but it often highlights too many textures, resulting in messy stylizations. Built upon WCT, Collab-Distill~\cite{wang2020collaborative} compresses the VGG-19 models, leading to better-stylized results. Nevertheless, the stylizations are still messy, with spurious boundaries ({\em e.g.}, the $2^{nd}$ row) and distorted contents ({\em e.g.}, the $3^{rd}$ row). In contrast, our MicroAST achieves very promising stylization effects. The contents are sharper and cleaner than WCT-U and Collab-Distill, while the colors and textures are more diverse and adequate than AdaIN-U and LST-U. Moreover, as shown in the bottom row, it can better preserve the content structures during style transfer, while other methods either lose the structural details or distort the content structures.

{\bf Quantitative Comparison.} Tab.~\ref{tab:quantity} shows the quantitative comparison with the state-of-the-art models. We collect 50 ultra-resolution (about 4K) content images and 40 ultra-resolution style images from \cite{wang2020collaborative,chen2022towards} and Internet to synthesize 2000 ultra-resolution results, and compute the average Structural Similarity Index (SSIM)~\cite{an2021artflow} and Style Loss~\cite{huang2017arbitrary} to assess the stylization quality in terms of content preservation and style transformation, respectively. As shown in columns (e) and (f), our method achieves the highest SSIM score and comparable Style Loss, indicating that it can transfer adequate style patterns while better preserving the content affinity. For efficiency (columns (a-d)), our MicroAST is 5-73 times smaller (column (a)) and 6-18 times faster (column (d)) than the state of the art, for the first time enabling {\em super-fast} AST on 4K ultra-resolution images.

{\bf User Study.} It is highly subjective to evaluate stylization results. Hence, we conducted a user study for the five approaches. We randomly showed each participant 30 septets of images consisting of the content, style, and five randomly shuffled outputs (AdaIN-U, LST-U, WCT-U, Collab-Distill, and ours). In each septet, they were given unlimited time to select their favorite output in terms of content preservation and stylization effects. We collect 1260 valid votes from 42 subjects and detail the preference percentage of each method in the last column of Tab.~\ref{tab:quantity}. The results demonstrate that our stylized images are more appealing than competitors.

\begin{figure}[t]
	\centering
	\includegraphics[width=1\columnwidth]{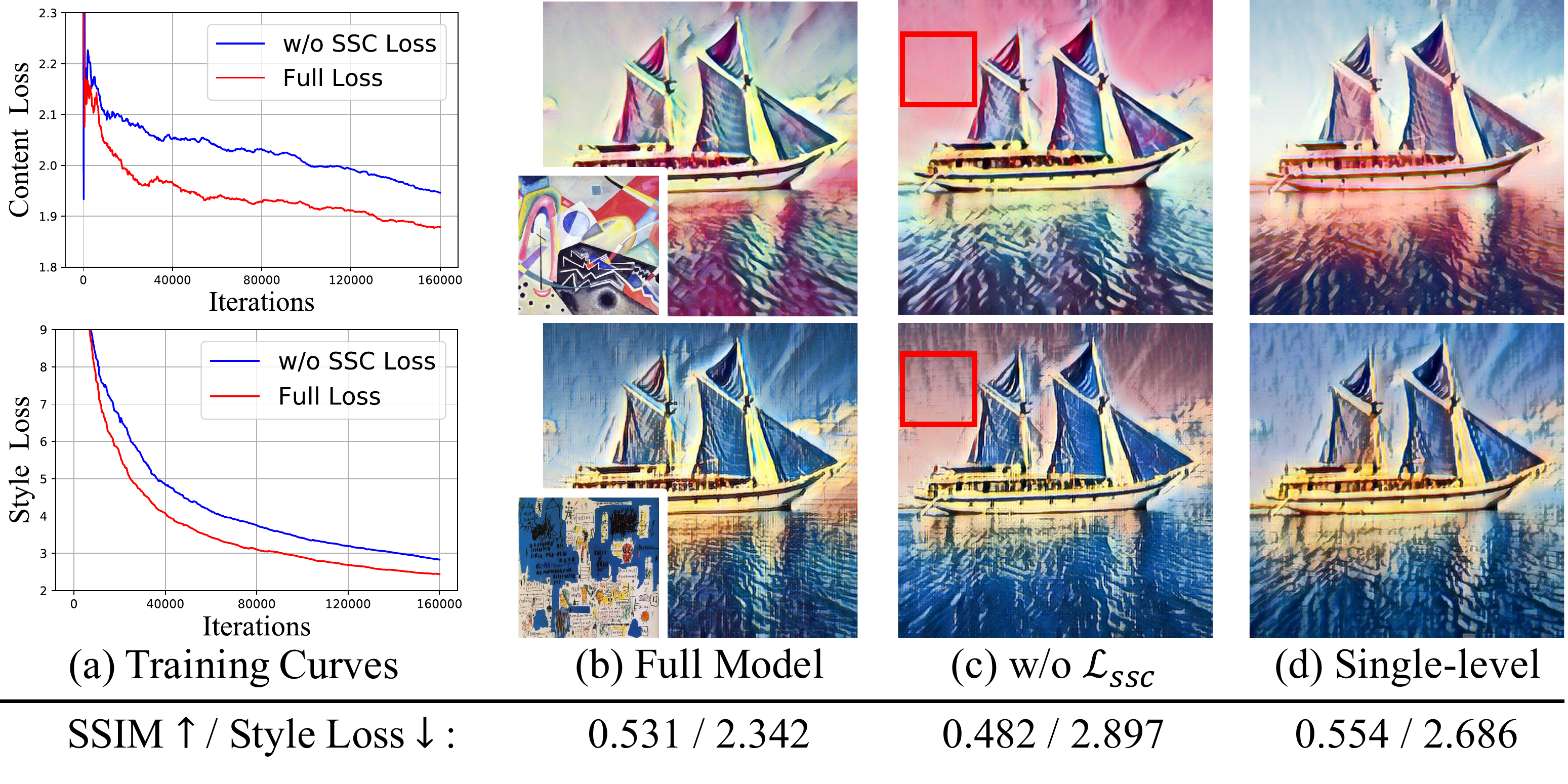}
	\caption{{\bf Ablation study} of (a,c) contrastive loss $\mathcal{L}_{ssc}$ and (d) single-level DualMod.
	}
	\label{fig:abs}
\end{figure}

%\vspace{-0.5em}
\subsection{Ablation Study}
\label{ablation}

{\bf With and Without Contrastive Loss $\mathcal{L}_{ssc}$.} We demonstrate the effect of our proposed style signal contrastive loss $\mathcal{L}_{ssc}$ in Fig.~\ref{fig:abs}. As shown in column (c), when training our MicroAST without $\mathcal{L}_{ssc}$, the stylization quality is significantly degraded where the colors from one style image may leak into the results stylized by other style images ({\em e.g.}, the pink color in the red box areas). It indicates that the micro style encoder $E_s$ is floundering in a compromised style representation that may map different styles to similar signals. This problem is alleviated after introducing $\mathcal{L}_{ssc}$ into training, which verifies that contrastive learning indeed helps $E_s$ to learn more distinct and representative style signals. Furthermore, it can also lead to faster and better convergence of content and style optimization and achieve higher SSIM score and lower Style Loss, as validated in column (a) and the bottom row of Fig.~\ref{fig:abs}. {\em More studies can be found in SM}.

{\bf Single-level DualMod vs. Multi-level DualMod.} We also compare the results of using single-level DualMod and multi-level DualMod in Fig.~\ref{fig:abs} (columns (d) and (b)). The multi-level design helps transfer more diverse colors and finer texture details, further improving stylization effects.

\section{Conclusion}
In this paper, we propose a straightforward and lightweight framework, dubbed MicroAST, for super-fast ultra-resolution arbitrary style transfer. A novel dual-modulation strategy is introduced to inject more sophisticated and flexible style signals to guide the stylizations. In addition, we also propose a new style signal contrastive loss to boost the ability of the style encoder to extract more distinct and representative style signals. Extensive experiments are conducted to demonstrate the effectiveness of our method. Compared to the state of the art, our MicroAST not only produces visually superior results but also is 5-73 times smaller and 6-18 times faster.

{\small
	\bibliographystyle{aaai23}
	\bibliography{aaai23}

\begin{thebibliography}{63}
\providecommand{\natexlab}[1]{#1}

\bibitem[{Alharbi, Smith, and Wonka(2019)}]{alharbi2019latent}
Alharbi, Y.; Smith, N.; and Wonka, P. 2019.
\newblock Latent filter scaling for multimodal unsupervised image-to-image
  translation.
\newblock In \emph{Proceedings of the IEEE Conference on Computer Vision and
  Pattern Recognition (CVPR)}, 1458--1466.

\bibitem[{An et~al.(2021)An, Huang, Song, Dou, Liu, and Luo}]{an2021artflow}
An, J.; Huang, S.; Song, Y.; Dou, D.; Liu, W.; and Luo, J. 2021.
\newblock ArtFlow: Unbiased Image Style Transfer via Reversible Neural Flows.
\newblock In \emph{Proceedings of the IEEE Conference on Computer Vision and
  Pattern Recognition (CVPR)}, 862--871.

\bibitem[{Champandard(2016)}]{champandard2016semantic}
Champandard, A.~J. 2016.
\newblock Semantic style transfer and turning two-bit doodles into fine
  artworks.
\newblock \emph{arXiv preprint arXiv:1603.01768}.

\bibitem[{Chandran et~al.(2021)Chandran, Zoss, Gotardo, Gross, and
  Bradley}]{chandran2021adaptive}
Chandran, P.; Zoss, G.; Gotardo, P.; Gross, M.; and Bradley, D. 2021.
\newblock Adaptive Convolutions for Structure-Aware Style Transfer.
\newblock In \emph{Proceedings of the IEEE Conference on Computer Vision and
  Pattern Recognition (CVPR)}, 7972--7981.

\bibitem[{Chen et~al.(2020{\natexlab{a}})Chen, Zhao, Qiu, Wang, Zhang, Xing,
  and Lu}]{chen2020creative}
Chen, H.; Zhao, L.; Qiu, L.; Wang, Z.; Zhang, H.; Xing, W.; and Lu, D.
  2020{\natexlab{a}}.
\newblock Creative and diverse artwork generation using adversarial networks.
\newblock \emph{IET Computer Vision}, 14(8): 650--657.

\bibitem[{Chen et~al.(2021{\natexlab{a}})Chen, Zhao, Wang, Zhang, Zuo, Li,
  Xing, and Lu}]{chen2021artistic}
Chen, H.; Zhao, L.; Wang, Z.; Zhang, H.; Zuo, Z.; Li, A.; Xing, W.; and Lu, D.
  2021{\natexlab{a}}.
\newblock Artistic Style Transfer with Internal-external Learning and
  Contrastive Learning.
\newblock In \emph{Advances in Neural Information Processing Systems
  (NeurIPS)}.

\bibitem[{Chen et~al.(2021{\natexlab{b}})Chen, Zhao, Wang, Zhang, Zuo, Li,
  Xing, and Lu}]{chen2021dualast}
Chen, H.; Zhao, L.; Wang, Z.; Zhang, H.; Zuo, Z.; Li, A.; Xing, W.; and Lu, D.
  2021{\natexlab{b}}.
\newblock DualAST: Dual Style-Learning Networks for Artistic Style Transfer.
\newblock In \emph{Proceedings of the IEEE Conference on Computer Vision and
  Pattern Recognition (CVPR)}, 872--881.

\bibitem[{Chen et~al.(2021{\natexlab{c}})Chen, Zhao, Zhang, Wang, Zuo, Li,
  Xing, and Lu}]{chen2021diverse}
Chen, H.; Zhao, L.; Zhang, H.; Wang, Z.; Zuo, Z.; Li, A.; Xing, W.; and Lu, D.
  2021{\natexlab{c}}.
\newblock Diverse image style transfer via invertible cross-space mapping.
\newblock In \emph{2021 IEEE/CVF International Conference on Computer Vision
  (ICCV)}, 14860--14869. IEEE Computer Society.

\bibitem[{Chen et~al.(2020{\natexlab{b}})Chen, Kornblith, Norouzi, and
  Hinton}]{chen2020simple}
Chen, T.; Kornblith, S.; Norouzi, M.; and Hinton, G. 2020{\natexlab{b}}.
\newblock A simple framework for contrastive learning of visual
  representations.
\newblock In \emph{International Conference on Machine Learning (ICML)},
  1597--1607. PMLR.

\bibitem[{Chen et~al.(2022)Chen, Wang, Xie, Lu, and Luo}]{chen2022towards}
Chen, Z.; Wang, W.; Xie, E.; Lu, T.; and Luo, P. 2022.
\newblock Towards Ultra-Resolution Neural Style Transfer via Thumbnail Instance
  Normalization.
\newblock In \emph{Proceedings of the AAAI Conference on Artificial
  Intelligence (AAAI)}, volume~36, 393--400.

\bibitem[{Cheng et~al.(2021)Cheng, Jaiswal, Wu, Natarajan, and
  Natarajan}]{cheng2021style}
Cheng, J.; Jaiswal, A.; Wu, Y.; Natarajan, P.; and Natarajan, P. 2021.
\newblock Style-Aware Normalized Loss for Improving Arbitrary Style Transfer.
\newblock In \emph{Proceedings of the IEEE Conference on Computer Vision and
  Pattern Recognition (CVPR)}, 134--143.

\bibitem[{Chiu(2019)}]{chiu2019understanding}
Chiu, T.-Y. 2019.
\newblock Understanding generalized whitening and coloring transform for
  universal style transfer.
\newblock In \emph{Proceedings of the IEEE/CVF International Conference on
  Computer Vision (ICCV)}, 4452--4460.

\bibitem[{Chiu and Gurari(2020)}]{chiu2020iterative}
Chiu, T.-Y.; and Gurari, D. 2020.
\newblock Iterative feature transformation for fast and versatile universal
  style transfer.
\newblock In \emph{Proceedings of the European Conference on Computer Vision
  (ECCV)}, 169--184. Springer.

\bibitem[{Deng et~al.(2021)Deng, Tang, Dong, Huang, Ma, and
  Xu}]{deng2021arbitrary}
Deng, Y.; Tang, F.; Dong, W.; Huang, H.; Ma, C.; and Xu, C. 2021.
\newblock Arbitrary video style transfer via multi-channel correlation.
\newblock In \emph{Proceedings of the AAAI Conference on Artificial
  Intelligence (AAAI)}, volume~35, 1210--1217.

\bibitem[{Deng et~al.(2022)Deng, Tang, Dong, Ma, Pan, Wang, and
  Xu}]{deng2022stytr2}
Deng, Y.; Tang, F.; Dong, W.; Ma, C.; Pan, X.; Wang, L.; and Xu, C. 2022.
\newblock StyTr2: Image Style Transfer with Transformers.
\newblock In \emph{Proceedings of the IEEE/CVF Conference on Computer Vision
  and Pattern Recognition (CVPR)}, 11326--11336.

\bibitem[{Deng et~al.(2020)Deng, Tang, Dong, Sun, Huang, and
  Xu}]{deng2020arbitrary}
Deng, Y.; Tang, F.; Dong, W.; Sun, W.; Huang, F.; and Xu, C. 2020.
\newblock Arbitrary style transfer via multi-adaptation network.
\newblock In \emph{Proceedings of the 28th ACM international conference on
  multimedia (ACM MM)}, 2719--2727.

\bibitem[{Gatys, Ecker, and Bethge(2016)}]{gatys2016image}
Gatys, L.~A.; Ecker, A.~S.; and Bethge, M. 2016.
\newblock Image style transfer using convolutional neural networks.
\newblock In \emph{Proceedings of the IEEE Conference on Computer Vision and
  Pattern Recognition (CVPR)}, 2414--2423.

\bibitem[{He et~al.(2020)He, Fan, Wu, Xie, and Girshick}]{he2020momentum}
He, K.; Fan, H.; Wu, Y.; Xie, S.; and Girshick, R. 2020.
\newblock Momentum contrast for unsupervised visual representation learning.
\newblock In \emph{Proceedings of the IEEE Conference on Computer Vision and
  Pattern Recognition (CVPR)}, 9729--9738.

\bibitem[{Hong et~al.(2021)Hong, Jeon, Yang, Fu, and Byun}]{hong2021domain}
Hong, K.; Jeon, S.; Yang, H.; Fu, J.; and Byun, H. 2021.
\newblock Domain-Aware Universal Style Transfer.
\newblock In \emph{Proceedings of the IEEE/CVF International Conference on
  Computer Vision (ICCV)}, 14609--14617.

\bibitem[{Howard et~al.(2017)Howard, Zhu, Chen, Kalenichenko, Wang, Weyand,
  Andreetto, and Adam}]{howard2017mobilenets}
Howard, A.~G.; Zhu, M.; Chen, B.; Kalenichenko, D.; Wang, W.; Weyand, T.;
  Andreetto, M.; and Adam, H. 2017.
\newblock Mobilenets: Efficient convolutional neural networks for mobile vision
  applications.
\newblock \emph{arXiv preprint arXiv:1704.04861}.

\bibitem[{Huang and Belongie(2017)}]{huang2017arbitrary}
Huang, X.; and Belongie, S. 2017.
\newblock Arbitrary style transfer in real-time with adaptive instance
  normalization.
\newblock In \emph{Proceedings of the IEEE International Conference on Computer
  Vision (ICCV)}, 1501--1510.

\bibitem[{Jing et~al.(2020)Jing, Liu, Ding, Wang, Ding, Song, and
  Wen}]{jing2020dynamic}
Jing, Y.; Liu, X.; Ding, Y.; Wang, X.; Ding, E.; Song, M.; and Wen, S. 2020.
\newblock Dynamic instance normalization for arbitrary style transfer.
\newblock In \emph{Proceedings of the AAAI Conference on Artificial
  Intelligence (AAAI)}, 4369--4376.

\bibitem[{Jing et~al.(2018)Jing, Liu, Yang, Feng, Yu, Tao, and
  Song}]{jing2018stroke}
Jing, Y.; Liu, Y.; Yang, Y.; Feng, Z.; Yu, Y.; Tao, D.; and Song, M. 2018.
\newblock Stroke controllable fast style transfer with adaptive receptive
  fields.
\newblock In \emph{Proceedings of the European Conference on Computer Vision
  (ECCV)}, 238--254.

\bibitem[{Jing et~al.(2022)Jing, Mao, Yang, Zhan, Song, Wang, and
  Tao}]{jing2022learning}
Jing, Y.; Mao, Y.; Yang, Y.; Zhan, Y.; Song, M.; Wang, X.; and Tao, D. 2022.
\newblock Learning Graph Neural Networks for Image Style Transfer.
\newblock In \emph{European Conference on Computer Vision (ECCV)}, 111--128.
  Springer.

\bibitem[{Jing et~al.(2019)Jing, Yang, Feng, Ye, Yu, and Song}]{jing2019neural}
Jing, Y.; Yang, Y.; Feng, Z.; Ye, J.; Yu, Y.; and Song, M. 2019.
\newblock Neural style transfer: A review.
\newblock \emph{IEEE Transactions on Visualization and Computer Graphics
  (TVCG)}, 26(11): 3365--3385.

\bibitem[{Johnson, Alahi, and Fei-Fei(2016)}]{johnson2016perceptual}
Johnson, J.; Alahi, A.; and Fei-Fei, L. 2016.
\newblock Perceptual losses for real-time style transfer and super-resolution.
\newblock In \emph{Proceedings of the European Conference on Computer Vision
  (ECCV)}, 694--711. Springer.

\bibitem[{Kang and Park(2020)}]{kang2020contragan}
Kang, M.; and Park, J. 2020.
\newblock ContraGAN: Contrastive Learning for Conditional Image Generation.
\newblock In \emph{Advances in Neural Information Processing Systems
  (NeurIPS)}.

\bibitem[{Karras, Laine, and Aila(2019)}]{karras2019style}
Karras, T.; Laine, S.; and Aila, T. 2019.
\newblock A style-based generator architecture for generative adversarial
  networks.
\newblock In \emph{Proceedings of the IEEE Conference on Computer Vision and
  Pattern Recognition (CVPR)}, 4401--4410.

\bibitem[{Karras et~al.(2020)Karras, Laine, Aittala, Hellsten, Lehtinen, and
  Aila}]{karras2020analyzing}
Karras, T.; Laine, S.; Aittala, M.; Hellsten, J.; Lehtinen, J.; and Aila, T.
  2020.
\newblock Analyzing and improving the image quality of stylegan.
\newblock In \emph{Proceedings of the IEEE Conference on Computer Vision and
  Pattern Recognition (CVPR)}, 8110--8119.

\bibitem[{Kingma and Ba(2015)}]{kingma2014adam}
Kingma, D.~P.; and Ba, J. 2015.
\newblock Adam: A method for stochastic optimization.
\newblock In \emph{International Conference on Learning Representations
  (ICLR)}.

\bibitem[{Kotovenko et~al.(2021)Kotovenko, Wright, Heimbrecht, and
  Ommer}]{kotovenko2021rethinking}
Kotovenko, D.; Wright, M.; Heimbrecht, A.; and Ommer, B. 2021.
\newblock Rethinking Style Transfer: From Pixels to Parameterized Brushstrokes.
\newblock In \emph{Proceedings of the IEEE Conference on Computer Vision and
  Pattern Recognition (CVPR)}, 12196--12205.

\bibitem[{Kwon and Ye(2022)}]{kwon2022clipstyler}
Kwon, G.; and Ye, J.~C. 2022.
\newblock Clipstyler: Image style transfer with a single text condition.
\newblock In \emph{Proceedings of the IEEE/CVF Conference on Computer Vision
  and Pattern Recognition (CVPR)}, 18062--18071.

\bibitem[{Li et~al.(2019)Li, Liu, Kautz, and Yang}]{li2019learning}
Li, X.; Liu, S.; Kautz, J.; and Yang, M.-H. 2019.
\newblock Learning linear transformations for fast image and video style
  transfer.
\newblock In \emph{Proceedings of the IEEE Conference on Computer Vision and
  Pattern Recognition (CVPR)}, 3809--3817.

\bibitem[{Li et~al.(2017)Li, Fang, Yang, Wang, Lu, and Yang}]{li2017universal}
Li, Y.; Fang, C.; Yang, J.; Wang, Z.; Lu, X.; and Yang, M.-H. 2017.
\newblock Universal style transfer via feature transforms.
\newblock In \emph{Advances in Neural Information Processing Systems
  (NeurIPS)}, 386--396.

\bibitem[{Lin et~al.(2021)Lin, Ma, Li, He, Li, Ding, Wang, Li, and
  Gao}]{lin2021drafting}
Lin, T.; Ma, Z.; Li, F.; He, D.; Li, X.; Ding, E.; Wang, N.; Li, J.; and Gao,
  X. 2021.
\newblock Drafting and Revision: Laplacian Pyramid Network for Fast
  High-Quality Artistic Style Transfer.
\newblock In \emph{Proceedings of the IEEE Conference on Computer Vision and
  Pattern Recognition (CVPR)}, 5141--5150.

\bibitem[{Lin et~al.(2014)Lin, Maire, Belongie, Hays, Perona, Ramanan,
  Doll{\'a}r, and Zitnick}]{lin2014microsoft}
Lin, T.-Y.; Maire, M.; Belongie, S.; Hays, J.; Perona, P.; Ramanan, D.;
  Doll{\'a}r, P.; and Zitnick, C.~L. 2014.
\newblock Microsoft coco: Common objects in context.
\newblock In \emph{Proceedings of the European Conference on Computer Vision
  (ECCV)}, 740--755. Springer.

\bibitem[{Liu et~al.(2021{\natexlab{a}})Liu, Ge, Choi, Wang, and
  Li}]{liu2021divco}
Liu, R.; Ge, Y.; Choi, C.~L.; Wang, X.; and Li, H. 2021{\natexlab{a}}.
\newblock Divco: Diverse conditional image synthesis via contrastive generative
  adversarial network.
\newblock In \emph{Proceedings of the IEEE Conference on Computer Vision and
  Pattern Recognition (CVPR)}, 16377--16386.

\bibitem[{Liu et~al.(2021{\natexlab{b}})Liu, Lin, He, Li, Wang, Li, Sun, Li,
  and Ding}]{liu2021adaattn}
Liu, S.; Lin, T.; He, D.; Li, F.; Wang, M.; Li, X.; Sun, Z.; Li, Q.; and Ding,
  E. 2021{\natexlab{b}}.
\newblock Adaattn: Revisit attention mechanism in arbitrary neural style
  transfer.
\newblock In \emph{Proceedings of the IEEE/CVF International Conference on
  Computer Vision (ICCV)}, 6649--6658.

\bibitem[{Lu et~al.(2019)Lu, Zhao, Yao, Chen, Xu, and Zhang}]{lu2019closed}
Lu, M.; Zhao, H.; Yao, A.; Chen, Y.; Xu, F.; and Zhang, L. 2019.
\newblock A Closed-Form Solution to Universal Style Transfer.
\newblock In \emph{Proceedings of the IEEE International Conference on Computer
  Vision (ICCV)}, 5952--5961.

\bibitem[{Oord, Li, and Vinyals(2018)}]{oord2018representation}
Oord, A. v.~d.; Li, Y.; and Vinyals, O. 2018.
\newblock Representation learning with contrastive predictive coding.
\newblock \emph{arXiv preprint arXiv:1807.03748}.

\bibitem[{Park and Lee(2019)}]{park2019arbitrary}
Park, D.~Y.; and Lee, K.~H. 2019.
\newblock Arbitrary style transfer with style-attentional networks.
\newblock In \emph{Proceedings of the IEEE Conference on Computer Vision and
  Pattern Recognition (CVPR)}, 5880--5888.

\bibitem[{Park et~al.(2020)Park, Efros, Zhang, and Zhu}]{park2020contrastive}
Park, T.; Efros, A.~A.; Zhang, R.; and Zhu, J.-Y. 2020.
\newblock Contrastive learning for unpaired image-to-image translation.
\newblock In \emph{Proceedings of the European Conference on Computer Vision
  (ECCV)}, 319--345. Springer.

\bibitem[{Phillips and Mackintosh(2011)}]{phillips2011wiki}
Phillips, F.; and Mackintosh, B. 2011.
\newblock Wiki Art Gallery, Inc.: A case for critical thinking.
\newblock \emph{Issues in Accounting Education}, 26(3): 593--608.

\bibitem[{Sanakoyeu et~al.(2018)Sanakoyeu, Kotovenko, Lang, and
  Ommer}]{sanakoyeu2018style}
Sanakoyeu, A.; Kotovenko, D.; Lang, S.; and Ommer, B. 2018.
\newblock A style-aware content loss for real-time hd style transfer.
\newblock In \emph{Proceedings of the European Conference on Computer Vision
  (ECCV)}, 698--714.

\bibitem[{Shen, Yan, and Zeng(2018)}]{shen2018neural}
Shen, F.; Yan, S.; and Zeng, G. 2018.
\newblock Neural style transfer via meta networks.
\newblock In \emph{Proceedings of the IEEE Conference on Computer Vision and
  Pattern Recognition (CVPR)}, 8061--8069.

\bibitem[{Sheng et~al.(2018)Sheng, Lin, Shao, and Wang}]{sheng2018avatar}
Sheng, L.; Lin, Z.; Shao, J.; and Wang, X. 2018.
\newblock Avatar-Net: Multi-scale Zero-shot Style Transfer by Feature
  Decoration.
\newblock In \emph{Proceedings of the IEEE Conference on Computer Vision and
  Pattern Recognition (CVPR)}, 8242--8250.

\bibitem[{Simonyan and Zisserman(2014)}]{simonyan2014very}
Simonyan, K.; and Zisserman, A. 2014.
\newblock Very deep convolutional networks for large-scale image recognition.
\newblock \emph{arXiv preprint arXiv:1409.1556}.

\bibitem[{Tian, Krishnan, and Isola(2020)}]{tian2020contrastive}
Tian, Y.; Krishnan, D.; and Isola, P. 2020.
\newblock Contrastive multiview coding.
\newblock In \emph{Proceedings of the European Conference on Computer Vision
  (ECCV)}, 776--794. Springer.

\bibitem[{Ulyanov et~al.(2016)Ulyanov, Lebedev, Vedaldi, and
  Lempitsky}]{ulyanov2016texture}
Ulyanov, D.; Lebedev, V.; Vedaldi, A.; and Lempitsky, V.~S. 2016.
\newblock Texture Networks: Feed-forward Synthesis of Textures and Stylized
  Images.
\newblock In \emph{International Conference on Machine Learning (ICML)},
  1349--1357.

\bibitem[{Wang et~al.(2020{\natexlab{a}})Wang, Li, Wang, Hu, and
  Yang}]{wang2020collaborative}
Wang, H.; Li, Y.; Wang, Y.; Hu, H.; and Yang, M.-H. 2020{\natexlab{a}}.
\newblock Collaborative Distillation for Ultra-Resolution Universal Style
  Transfer.
\newblock In \emph{Proceedings of the IEEE Conference on Computer Vision and
  Pattern Recognition (CVPR)}, 1860--1869.

\bibitem[{Wang et~al.(2022{\natexlab{a}})Wang, Zhang, Zhao, Zuo, Li, Xing, and
  Lu}]{wang2022aesust}
Wang, Z.; Zhang, Z.; Zhao, L.; Zuo, Z.; Li, A.; Xing, W.; and Lu, D.
  2022{\natexlab{a}}.
\newblock AesUST: towards aesthetic-enhanced universal style transfer.
\newblock In \emph{Proceedings of the 30th ACM International Conference on
  Multimedia (ACM MM)}, 1095--1106.

\bibitem[{Wang et~al.(2022{\natexlab{b}})Wang, Zhao, Chen, Li, Zuo, Xing, and
  Lu}]{wang2022texture}
Wang, Z.; Zhao, L.; Chen, H.; Li, A.; Zuo, Z.; Xing, W.; and Lu, D.
  2022{\natexlab{b}}.
\newblock Texture reformer: Towards fast and universal interactive texture
  transfer.
\newblock In \emph{Proceedings of the AAAI Conference on Artificial
  Intelligence (AAAI)}, volume~36, 2624--2632.

\bibitem[{Wang et~al.(2020{\natexlab{b}})Wang, Zhao, Chen, Qiu, Mo, Lin, Xing,
  and Lu}]{wang2020diversified}
Wang, Z.; Zhao, L.; Chen, H.; Qiu, L.; Mo, Q.; Lin, S.; Xing, W.; and Lu, D.
  2020{\natexlab{b}}.
\newblock Diversified Arbitrary Style Transfer via Deep Feature Perturbation.
\newblock In \emph{Proceedings of the IEEE Conference on Computer Vision and
  Pattern Recognition (CVPR)}, 7789--7798.

\bibitem[{Wang et~al.(2021)Wang, Zhao, Chen, Zuo, Li, Xing, and
  Lu}]{wang2021evaluate}
Wang, Z.; Zhao, L.; Chen, H.; Zuo, Z.; Li, A.; Xing, W.; and Lu, D. 2021.
\newblock Evaluate and improve the quality of neural style transfer.
\newblock \emph{Computer Vision and Image Understanding (CVIU)}, 207: 103203.

\bibitem[{Wang et~al.(2022{\natexlab{c}})Wang, Zhao, Chen, Zuo, Li, Xing, and
  Lu}]{wang2022divswapper}
Wang, Z.; Zhao, L.; Chen, H.; Zuo, Z.; Li, A.; Xing, W.; and Lu, D.
  2022{\natexlab{c}}.
\newblock DivSwapper: towards diversified patch-based arbitrary style transfer.
\newblock In \emph{Proc. Int. Joint Conf. on Artif. Intell.(IJCAI)}, volume~36,
  4980--4987.

\bibitem[{Wang et~al.(2020{\natexlab{c}})Wang, Zhao, Lin, Mo, Zhang, Xing, and
  Lu}]{wang2020glstylenet}
Wang, Z.; Zhao, L.; Lin, S.; Mo, Q.; Zhang, H.; Xing, W.; and Lu, D.
  2020{\natexlab{c}}.
\newblock GLStyleNet: exquisite style transfer combining global and local
  pyramid features.
\newblock \emph{IET Computer Vision}, 14(8): 575--586.

\bibitem[{Wu et~al.(2021)Wu, Qu, Lin, Zhou, Qiao, Zhang, Xie, and
  Ma}]{wu2021contrastive}
Wu, H.; Qu, Y.; Lin, S.; Zhou, J.; Qiao, R.; Zhang, Z.; Xie, Y.; and Ma, L.
  2021.
\newblock Contrastive Learning for Compact Single Image Dehazing.
\newblock In \emph{Proceedings of the IEEE Conference on Computer Vision and
  Pattern Recognition (CVPR)}, 10551--10560.

\bibitem[{Wu et~al.(2022)Wu, Zhu, Du, and Bai}]{wu2022ccpl}
Wu, Z.; Zhu, Z.; Du, J.; and Bai, X. 2022.
\newblock CCPL: Contrastive Coherence Preserving Loss for Versatile Style
  Transfer.
\newblock In \emph{European Conference on Computer Vision (ECCV)}, 189--206.
  Springer.

\bibitem[{Xie et~al.(2022)Xie, Li, Huang, Fu, Wang, and Guo}]{xie2022artistic}
Xie, X.; Li, Y.; Huang, H.; Fu, H.; Wang, W.; and Guo, Y. 2022.
\newblock Artistic Style Discovery With Independent Components.
\newblock In \emph{Proceedings of the IEEE/CVF Conference on Computer Vision
  and Pattern Recognition (CVPR)}, 19870--19879.

\bibitem[{Zhang, Zhu, and Zhu(2019)}]{zhang2019metastyle}
Zhang, C.; Zhu, Y.; and Zhu, S.-C. 2019.
\newblock MetaStyle: Three-Way Trade-off among Speed, Flexibility, and Quality
  in Neural Style Transfer.
\newblock In \emph{Proceedings of the AAAI Conference on Artificial
  Intelligence (AAAI)}, volume~33, 1254--1261.

\bibitem[{Zhang et~al.(2022{\natexlab{a}})Zhang, Li, Li, Jia, and
  Zhang}]{zhang2022exact}
Zhang, Y.; Li, M.; Li, R.; Jia, K.; and Zhang, L. 2022{\natexlab{a}}.
\newblock Exact feature distribution matching for arbitrary style transfer and
  domain generalization.
\newblock In \emph{Proceedings of the IEEE/CVF Conference on Computer Vision
  and Pattern Recognition (CVPR)}, 8035--8045.

\bibitem[{Zhang et~al.(2022{\natexlab{b}})Zhang, Tang, Dong, Huang, Ma, Lee,
  and Xu}]{zhang2022domain}
Zhang, Y.; Tang, F.; Dong, W.; Huang, H.; Ma, C.; Lee, T.-Y.; and Xu, C.
  2022{\natexlab{b}}.
\newblock Domain Enhanced Arbitrary Image Style Transfer via Contrastive
  Learning.
\newblock \emph{arXiv preprint arXiv:2205.09542}.

\bibitem[{Zuo et~al.(2022)Zuo, Zhao, Lian, Chen, Wang, Li, Xing, and
  Lu}]{zuo2022style}
Zuo, Z.; Zhao, L.; Lian, S.; Chen, H.; Wang, Z.; Li, A.; Xing, W.; and Lu, D.
  2022.
\newblock Style fader generative adversarial networks for style degree
  controllable artistic style transfer.
\newblock In \emph{Proc. Int. Joint Conf. on Artif. Intell.(IJCAI)},
  5002--5009.

\end{thebibliography}
}

\end{document}